\documentclass[11pt]{article}

\usepackage[preprint]{acl}

\usepackage{times}
\usepackage{latexsym}

\usepackage[T1]{fontenc}

\usepackage[utf8]{inputenc}

\usepackage{microtype}

\usepackage{inconsolata}

\usepackage{graphicx}
\usepackage{booktabs}
\usepackage{multicol}
\usepackage{multirow}
\usepackage{verbatim}
\usepackage{subfigure}
\usepackage{subcaption}

\title{Which Tokens Need Context? A Reference-Based Analysis of Translation Responsibility Using Fertility and Entropy}

\author{Ramakrishna Appicharla$^1$, Baban Gain$^1$, Santanu Pal$^2$, Asif Ekbal$^1$ \\ $^1$Department of Computer Science and Engineering, Indian Institute of Technology Patna, India \\ $^2$Wipro AI Lab45, Bengaluru, India \\ {\tt \{ramakrishnaappicharla,gainbaban, santanu.pal.ju, asif.ekbal\} @gmail.com}}

\begin{document}
\maketitle
\begin{abstract}
When humans translate, not every word depends equally on the surrounding context. Some tokens, particularly function words like pronouns and auxiliaries, rely heavily on preceding or following sentences, while others, such as proper nouns, do not. Understanding this inherent context sensitivity is essential for evaluating whether machine translation systems use context in human-like ways. However, existing approaches to analysing context usage rely on discourse-specific test sets or model internals, making them narrow or model-dependent. We propose a post-hoc, model-agnostic framework to quantify context sensitivity at lexical and syntactic levels using two measures derived from word alignments: fertility (number of target tokens generated per source token) and entropy (stability of fertility patterns across contexts). Using reference translations for three language pairs (German $\leftrightarrow$ English, English $\rightarrow$ Hindi) under four context conditions, we show that context selectively redistributes generative responsibility from source to context tokens without altering overall fertility. Function words show the largest fertility reductions, while content words remain stable, suggesting that context resolves ambiguity rather than adding new information. Our framework provides a ground-truth characterisation of selective context usage in human translation, establishing a diagnostic baseline for evaluating machine translation models.
\end{abstract}

\section{Introduction}
\label{sec:introduction}
Extra sentential context plays a major role in understanding ambiguous phenomena such as anaphora, ellipses, and other discourse properties. Humans extract specific linguistic cues from the context to resolve such ambiguities. This is a well-established finding in psycholinguistics \citep{malt1985role,van1999context,van2004sentence} that humans process context selectively rather than uniformly. This selective, non-linear context processing is a fundamental property of human language processing.

However, the current computational models, particularly document-level machine translation (DocMT) systems encode context densely and sequentially by treating all the tokens as potentially relevant \citep{voita-etal-2018-context,maruf-haffari-2018-document,zhang-etal-2018-improving,agrawal-etal-2018-contextual,lopes-etal-2020-document,donato-etal-2021-diverse}. Yet, several studies have reported that context is not utilised in an interpretable way \citep{kim-etal-2019-document,kim-2025-context,castilho2025survey} and even randomly sampled context also improves the translation performance \citep{li-etal-2020-multi-encoder,bawden-yvon-2023-investigating,appicharla-etal-2023-case}. This suggests that the models may be either ignoring the context or learning spurious correlations between the source and the context. Hence, it is essential to study how the extra sentential context influences the source and target in translation settings.


Broadly, the prior research has approached the analysis of context utilisation in two directions. The first direction is to evaluate the DocMT models on discourse-specific test sets \citep{bawden2021diabla,voita-etal-2019-good,fernandes-etal-2023-translation,futeral-etal-2025-moscar} and contrastive test sets \citep{muller-etal-2018-large,bawden-etal-2018-evaluating,yin-etal-2021-context,post2023escaping}. The second direction is to analyse the model behaviour through explainable AI (XAI) techniques such as mutual information analysis \citep{fernandes-etal-2021-measuring,fernandes-etal-2023-translation}, attention visualisation \citep{sarti2023quantifying,maka-etal-2025-analyzing}, perturbation and attribution analysis \citep{mohammed-niculae-2024-measuring,mohammed-niculae-2025-context}. Although these evaluations provide precise insights into the models' behaviour, they are limited to some specific discourse phenomena and require specifically constructed datasets as well as model internal representations.

In this work, we present a complementary, post-hoc framework that does not rely on specific discourse test sets or model internal representations. Instead, we quantify selective context utilisation at lexical and syntactic levels using two measures derived from word alignments: fertility and entropy. Fertility \citep{brown-etal-1993-mathematics} measures the number of target tokens generated from a source token and captures how much structural contribution a token makes to the translation. Entropy measures the stability of fertility patterns across contexts, capturing how consistently tokens participate in translation. Together, these two measures allow us to systematically study how generative responsibility is redistributed between source and context tokens when context is introduced.



We apply this framework to reference translations, i.e., human translations, as a case study. Reference translations provide a natural distribution of discourse phenomena, whereas contrastive test sets generally focus on a specific phenomenon. The analysis is performed on reference translations rather than model outputs, allowing us to capture underlying structural patterns instead of model-specific generation behaviour. By analysing which tokens show fertility reductions when context is added, and which tokens exhibit entropy shifts, we can identify which linguistic categories are genuinely context-sensitive. Specifically, we ask: (1) Which tokens require context?, (2) Does context redistribute responsibility or add new information?, and (3) How consistent is this redistribution across languages and context types?


The contributions of our work can be summarised as follows:

\begin{itemize}
    \item We introduce a post-hoc, model-agnostic methodology for quantifying selective context usage using fertility and entropy derived from word alignments.
    
    \item We propose the notion of responsibility redistribution: when context is added, generative responsibility shifts from source tokens to context tokens without increasing overall fertility, indicating selective rather than additive context use.
    
    \item We provide a systematic empirical analysis across three language pairs (German $\leftrightarrow$ English, English $\rightarrow$ Hindi) and four context settings (no context, previous sentence, next sentence, random sentence), showing that function words (pronouns, auxiliaries, particles, conjunctions) are the primary targets of selective context attention.
    
    \item We show that fertility and entropy together offer interpretable signals for understanding where contextual information contributes to translation and how consistently it is utilised, establishing a diagnostic baseline for future evaluation of machine translation systems.
\end{itemize}

\section{Related Work}
\subsection{Document-level Machine Translation}
The main goal of document-level machine translation is to improve the accuracy of discourse-level phenomena and coherence in translation \citep{maruf2021survey}. There have been many different approaches proposed for encoding context with traditional transformer-based models \citep{tiedemann-scherrer-2017-neural,agrawal2018contextual,zhang-etal-2018-improving,voita-etal-2018-context,ma-etal-2020-simple} and Large Language Model (LLM)-based \citep{zhang-etal-2023-machine,wang2024delta,wu2024adapting} approaches. Similarly, various metrics \citep{miculicich-werlen-popescu-belis-2017-validation,liu-etal-2020-multilingual-denoising,bao-etal-2021-g,lyu-etal-2021-encouraging,jiang-etal-2022-blonde,vernikos-etal-2022-embarrassingly} have also been proposed to evaluate the performance of DocMT systems. While these works focus on improving DocMT systems and measuring the overall performance, our work focuses on analysing context utilisation.

\subsection{Analysing Context Utilisation}
Broadly, there are two research directions for analysing the utilisation of context: using contrastive test sets and via explainable AI techniques. Contrastive test sets are designed to evaluate DocMT models' ability to discriminate between correct and incorrect translations for specific discourse phenomena. Models generally perform better on these contrastive test sets when the context is sufficiently encoded. \citet{muller-etal-2018-large} built a pronoun translation test set for the German-English language pair. \citet{lopes-etal-2020-document} extended the dataset created by \citet{muller-etal-2018-large} for the English-French language pair. \citet{bawden-etal-2018-evaluating} created an English-French dataset for pronoun translation and lexical cohesion. \citet{voita-etal-2019-good} built an English-Russian dataset for deixis, ellipsis and lexical cohesion. These datasets are discriminative in nature, and \citet{post2023escaping} later created generative variants of them.

Explainable AI (XAI) approaches aim to probe a model's internal representations to better understand the reasoning behind its predictions. \citet{fernandes-etal-2021-measuring} proposed the conditional cross-mutual information (C-XMI) metric, which measures the difference between the cross-entropies of context-agnostic and context-aware systems. This measure provides how much information the context provides, given a source-target pair. They further extend this approach to create a benchmark for evaluating discourse phenomena \citep{fernandes-etal-2023-translation} for 14 language pairs. \citet{sarti2023quantifying} proposed an end-to-end interpretability framework to identify context-sensitive tokens and link them to contextual cues. \citet{mohammed-niculae-2024-measuring} performed perturbation and attribution analyses to analyse the effects of correct vs. random context and context tokens that influence pronoun translation performance. Further, they extended their framework to LLM-based DocMT models \citep{mohammed-niculae-2025-context}. Similarly, \citet{maka-etal-2025-analyzing} analysed the role of attention heads in context-aware MT settings. In contrast to prior work, our work analyses the context utilisation at the lexical and syntactic levels.

\subsection{Word Alignment-based Analysis in Machine Translation}
There have been some works to predict and measure translation difficulty using word alignments. \citet{lim-etal-2023-predicting} computed entropy and surprisal \citep{wei-2022-entropy} values based on word alignments, and correlated them with the behavioural dataset \citep{carl2016critt}. They report that word alignment-based measures can effectively estimate translation difficulty. Further, they extend the alignment-based measures to attention-based measures \citep{lim-etal-2024-predicting}. 

\citet{brown-etal-1993-mathematics} introduced the concept of fertility in the IBM model 3 of word-based statistical machine translation systems, which estimates the number of target words a particular source word generates. In deep learning models, fertility is used in non-autoregressive translation models \citep{gu2017non,xiao2023survey,song-etal-2021-alignart} to estimate the length of the target sequence given a source sequence. In this work, instead of directly using the word alignment information, we calculate fertility to systematically attribute the responsibility of generating target tokens to source and context tokens.

\section{Methodology}
Our goal is to analyse context utilisation without probing model internal representations or discourse-specific test sets. To this end, we adopt a post-hoc analysis framework based on word alignment information. Based on the alignment statistics, we study which input (source and context) tokens are responsible for generating target tokens at the lexical and syntactic levels via fertility and entropy measures.

\subsection{Fertility Computation}
\label{sec:fertility_computation}
Given a dataset $\mathcal{D}$ consisting of source–target sentence pairs, $\mathcal{S}=\{s_1,s_2,\dots,s_n\}$ and $\mathcal{T}=\{t_1,t_2,\dots,t_n\}$, we first extract token-level alignments $\mathcal{A}=\{a_1,a_2,\dots,a_n\}$ using the \texttt{awesome-align} toolkit \citep{dou-neubig-2021-word}. We use the provided fine-tuned multilingual model without the \texttt{-{}-train\_co} option.\footnote{Alignment extraction settings follow the default inference configuration of the toolkit.} Each element $a_i$ denotes the set of alignment links between the source sentence $s_i$ and the target sentence $t_i$.

We use the extracted alignments to compute fertility, defined as the number of target tokens aligned to a given input token \citep{brown-etal-1993-mathematics}. Specifically, let the source sentence $s_i=\{s_i^1,s_i^2,\dots,s_i^k\}$ contain $k$ tokens and the target sentence $t_i=\{t_i^1,t_i^2,\dots,t_i^m\}$ contain $m$ tokens. The fertility of a source token $s_i^j$ is defined as:

\begin{equation}
    f(s_i^j)\,=\,|\{t\in t_i\,|\,(s_i^j,t)\in a_i\}|
    \label{eq:fertility}
\end{equation}

The average sentence fertility $f_i$ is then computed as:

\begin{equation}
    f_i\,=\,\frac{1}{k}\sum_{j=1}^kf(s_i^j)
    \label{eq:avg_sent_fertility}
\end{equation}

which corresponds to the average number of target tokens generated per source token. We counted multiple alignment links to the same target token independently, since multiple source tokens may contribute to the generation of a single target token\footnote{We compute the sentence-level fertility only over source tokens that have at least one alignment link, and tokens with zero fertility are analysed separately through the zero-fertility coverage statistics (cf. Section \ref{sec:source_zero_fertility_analysis})}. The corpus-level fertility is obtained by averaging sentence-level fertility across the dataset as:

\begin{equation}
    f_{\mathcal{D}}\,=\,\frac{1}{n}\sum_{i=1}^nf_i
    \label{eq:avg_corpus_fertility}
\end{equation}

To analyse token-level fertility patterns, we group tokens according to their Part-of-Speech (PoS) categories and compute the average fertility within each category. PoS tags provide a linguistically meaningful way to group tokens and enable comparison of fertility patterns across different language pairs. We use the \texttt{Stanza} library \cite{qi-etal-2020-stanza} to annotate the source tokens with Universal PoS (UPOS) tags.

Let $pos_c$ denote a PoS category. Tokens belonging to $pos_c$ may exhibit multiple fertility values, including zero-fertility tokens (unaligned tokens). If $n_{pos_c}$ denotes the total number of tokens in category $pos_c$, and $|s_{c_y}|$ denotes the number of tokens with fertility value $y$, the average fertility for the category is computed as:

\begin{equation}
    f_{pos_c}\,=\,\frac{1}{n_{pos_c}}\sum_{y=0}^Fy\,*\,|s_{c_y}|
    \label{eq:pos_avg_fertility}
\end{equation}

where $F$ denotes distinct fertility categories that a given PoS tag takes.

A token with fertility greater than zero contributes explicitly to the generation of one or more target tokens, while tokens with zero fertility do not produce directly aligned target tokens. When contextual information is introduced, shifts in fertility patterns can reveal how translation responsibility is redistributed between the source and context tokens.

\subsection{Entropy of Fertility Distributions}
\label{sec:entropy_of_fertility}
While fertility identifies which tokens (from both source and context) contribute to the generation of target tokens, it does not capture how consistently this responsibility is distributed across multiple translation instances. Tokens belonging to the same part-of-speech (PoS) category may exhibit different fertility patterns depending on the surrounding context. To measure this variability, we compute the entropy of the fertility distribution for each PoS category.

Since entropy is defined over probability distributions, we first construct a distribution of fertility values for each PoS category. Let $pos_c$ denote a PoS category. For tokens belonging to $pos_c$, we compute the probability of observing fertility value $y$ as:

\begin{equation}
    p(pos_{c_y})\,=\,\frac{|s_{c_y}|}{n_{pos_c}}
    \label{eq:pos_entropy_probability}
\end{equation}

where $|s_{c_y}|$ denotes the number of tokens in category $pos_c$ with fertility value $y$, and $n_{pos_c}$ denotes the total number of tokens belonging to that PoS category. The entropy of the fertility distribution for category $pos_c$ is then computed using Shannon entropy as:

\begin{equation}
    \mathcal{H}\,=-\,\sum_{y=0}^F\,p(pos_{c_y})\log p(pos_{c_y})
    \label{eq:pos_entropy}
\end{equation}

where $F$ denotes the maximum observed fertility value for the category. To facilitate comparison across PoS categories, we normalise the entropy by dividing it by $\log(F)$.

Low entropy indicates that tokens from a given category tend to exhibit stable fertility patterns, whereas higher entropy suggests greater variability in how those tokens participate in translation.

We calculate fertility and entropy values on source and context sentences independently, as well as on the combined (source + context) inputs, to clearly examine how the patterns vary between source and context sentences. Together, fertility and entropy provide complementary perspectives on context utilisation. Specifically, increases in the fertility of context tokens accompanied by corresponding shifts in the source fertility values may indicate redistribution of responsibility between these two types.

\section{Experimental Setup}
\subsection{Datasets and Preprocessing}
We conduct experiments on German-to-English, English-to-German, and English-to-Hindi language directions. We utilise IWSLT`17 TED \citep{maruf-etal-2019-selective}\footnote{\url{https://github.com/sameenmaruf/selective-attn/tree/master/data}} and IN-22 \citep{gala2023indictrans}\footnote{\url{https://github.com/AI4Bharat/IndicTrans2}} test sets for German-English and Hindi-English language pairs, respectively. We tokenise the data with \texttt{sacremoses}\footnote{\url{https://github.com/hplt-project/sacremoses}} tokeniser for English and German, and \texttt{Indic-NLP-Library}\footnote{\url{https://github.com/anoopkunchukuttan/indic_nlp_library}} for Hindi languages. Word alignments are extracted using \texttt{awesome-align} \citep{dou-neubig-2021-word} toolkit\footnote{\url{https://github.com/neulab/awesome-align}}. We use the same tokenised data while extracting PoS tags with \texttt{Stanza}\footnote{\url{https://stanfordnlp.github.io/stanza/index.html}} library\footnote{We set \texttt{tokenize\_pretokenized=True} in the Stanza PoS tagging pipeline}. Table \ref{tab:data_statistics} shows the statistics of the datasets used in the experiments.

\begin{table}[!ht]
    \centering
    \resizebox{1.0\linewidth}{!}{
    \begin{tabular}{cccc}
        \toprule
        \textbf{Language Pair} & \textbf{Domain} & \textbf{\# Docs} & \textbf{\# Sents} \\
        \midrule
        De-En & TED & 23 & 2,271 \\
        Hi-En & Conversational & 44 & 1,503 \\
        \bottomrule
    \end{tabular}
    }
    \caption{Statistics of data used in the experiments. \textbf{\# Docs} and \textbf{\# Sents} represent number of documents and sentences, respectively.}
    \label{tab:data_statistics}
\end{table}

\subsection{Context Selection}
We utilise the context from the source side, and the context is prepended\footnote{We do not use any tags while concatenating the context and source sentences.} to the source sentence. We primarily conduct experiments on four different context settings:

\begin{itemize}
    \item No Context: The baseline setting where no context is given.
    \item Previous Context (Prev-1): One previous sentence to the current source sentence.
    \item Next Context (Next-1): One next sentence to the current source sentence.
    \item Random Context (Random-1): One randomly sampled sentence across all documents.
\end{itemize}

In the random context setting, each experiment is repeated two times, and the averages of the obtained statistics are used for analysis.

\section{Results and Analysis}
\subsection{Corpus-Level Fertility Redistribution}
\begin{figure*}[!ht]
    \centering
    \subfigure[]{\includegraphics[width=0.32\textwidth]{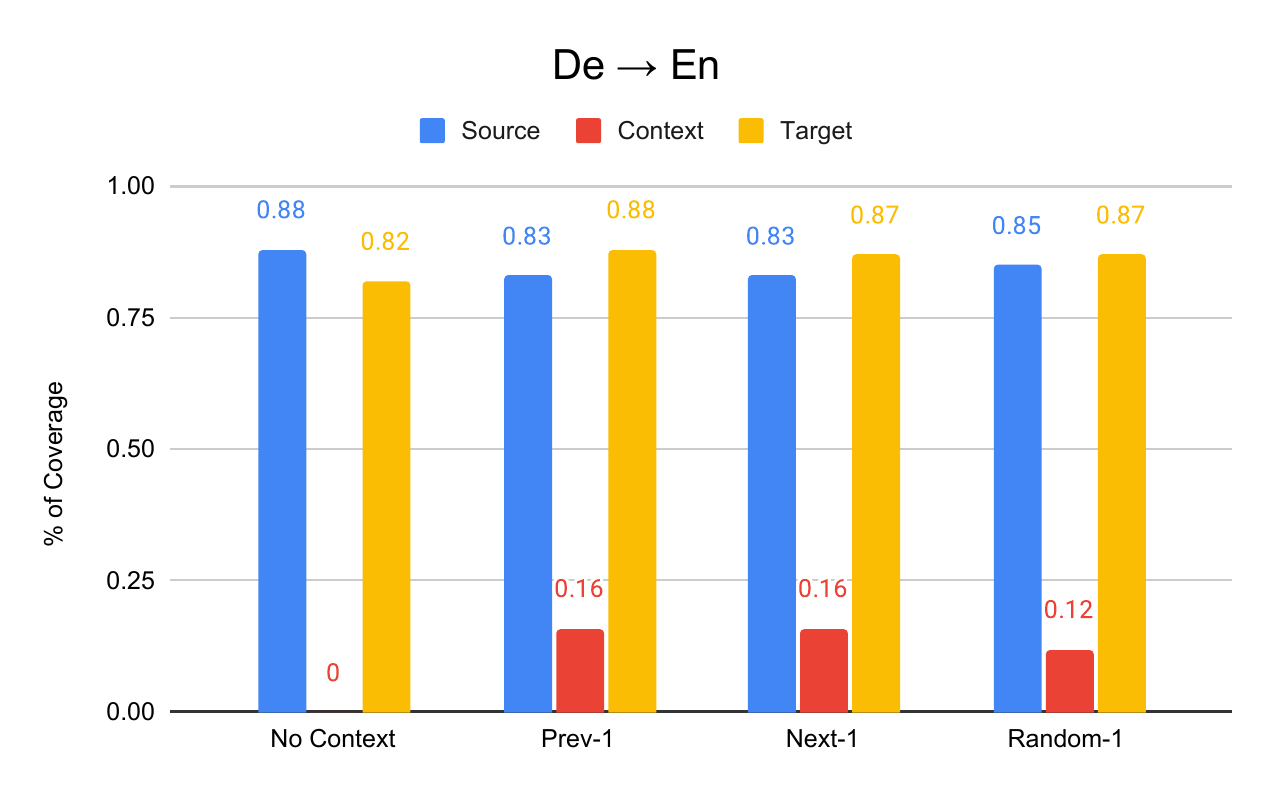}}
    \subfigure[]{\includegraphics[width=0.32\textwidth]{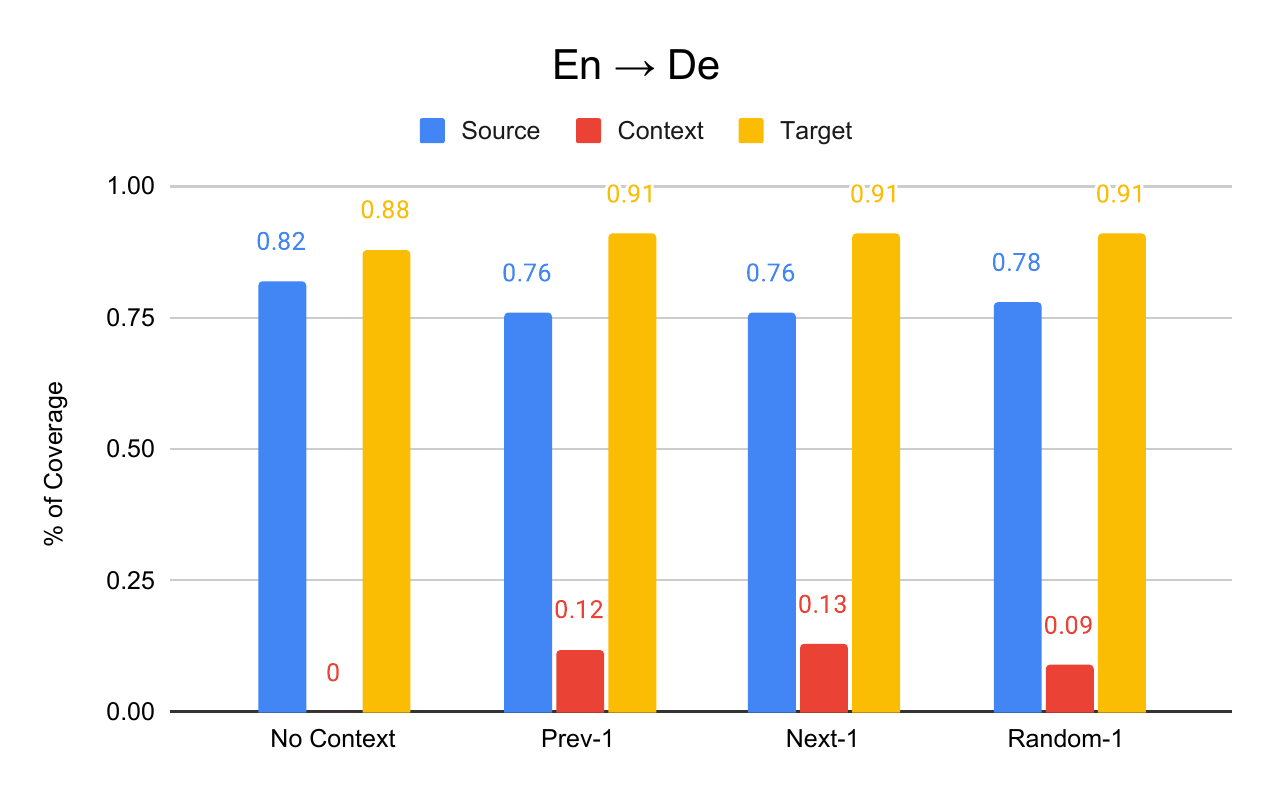}}
    \subfigure[]{\includegraphics[width=0.32\textwidth]{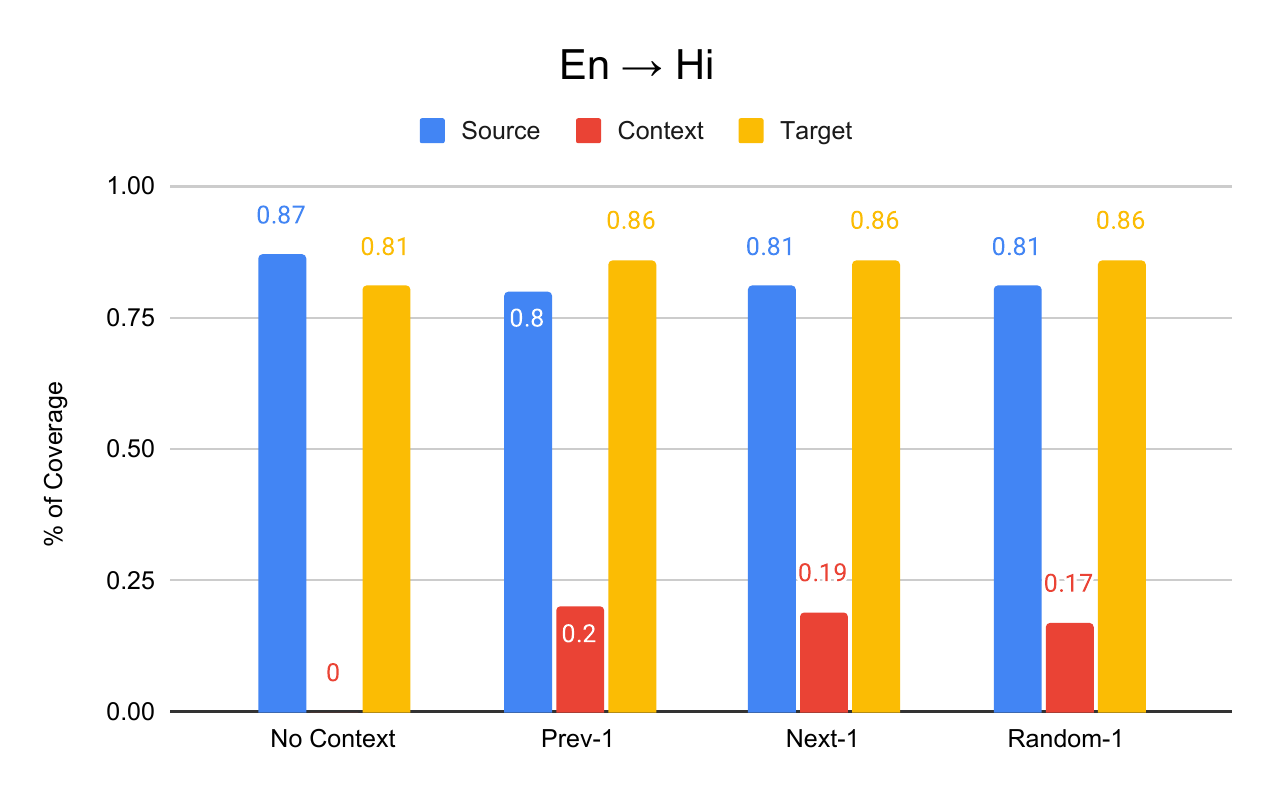}}
    \caption{Coverage analysis of the source, context, and target corpora for different context settings. Figures (a), (b), and (c) represent coverage statistics for De $\rightarrow$ En, En $\rightarrow$ De, and En $\rightarrow$ Hi language pairs, respectively. The Y-axis shows the percentage of tokens having at least one alignment link between source (or context) and target tokens.}
    \label{fig:sentence_coverage}
\end{figure*}

Table \ref{tab:avg_corpus_fertility} shows the average corpus-level fertility (cf. Equation \ref{eq:avg_corpus_fertility}) for different context settings. We observe that adding context does not affect the overall fertility of the corpus, as the average fertility values are almost identical for different context settings. This observation is consistent with previous studies, which report that adding context often leads to only modest improvements in BLEU scores \citep{li-etal-2020-multi-encoder,bawden-yvon-2023-investigating}. However, we hypothesise that adding context may shift some of the generative responsibility from source tokens to context tokens. To study this, we plot the percentage of tokens that have at least one alignment link (fertility value $> 0$) from source, context, and target sentences. Figure \ref{fig:sentence_coverage} shows source, context, and target coverage for different context settings.

\begin{table}[!ht]
    \centering
    \resizebox{1.0\linewidth}{!}{
    \begin{tabular}{ccccc}
        \toprule
        \textbf{Lang Pair} & \textbf{No Context} & \textbf{Prev-1} & \textbf{Next-1} & \textbf{Random-1} \\
        \midrule
        De $\rightarrow$ En & 1.05 & 1.05 & 1.05 & 1.05 \\
        \cmidrule{1-5}
        En $\rightarrow$ De & 1.02 & 1.01 & 1.01 & 1.01 \\
        \cmidrule{1-5}
        En $\rightarrow$ Hi & 1.07 & 1.06 & 1.06 & 1.06 \\
        \bottomrule
    \end{tabular}
    }
    \caption{Average corpus-level fertility with different context settings.}
    \label{tab:avg_corpus_fertility}
\end{table}

Although the average fertility does not change, we can clearly observe that adding context reduces source coverage by 5–7\% while introducing 9–20\% coverage from context tokens. Target token coverage also increases by 3–6\%. This highlights that adding context takes away some generative responsibility from source tokens and redistributes it to context tokens, which results in a higher proportion of target tokens being aligned. Interestingly, random context also produces non-zero alignment coverage. This is expected because alignment models may link semantically or lexically similar tokens even when the context sentence is unrelated. However, there is also an increase in source token coverage compared to the \texttt{Prev-1} and \texttt{Next-1} settings when context is random. Overall, the results suggest that adding context does not change the overall fertility but redistributes the responsibility for generating target tokens. We further study how PoS categories behave when context is added.

\subsection{Source PoS Fertility Analysis}
\label{sec:source_pos_fertility_analysis}

\begin{table*}[!ht]
    \centering
    \resizebox{1.0\linewidth}{!}{
    \begin{tabular}{l||c|cccc||c|cccc||c|ccc}
         \toprule
         \multirow{2}*{\textbf{PoS Tag}} & \multicolumn{4}{c}{\textbf{De $\rightarrow$ En}} & & \multicolumn{4}{c}{\textbf{En $\rightarrow$ De}} & & \multicolumn{4}{c}{\textbf{En $\rightarrow$ Hi}} \\
         \cmidrule{2-15}
          & No Context & $\Delta f_{Prev-1}$ & $\Delta f_{Next-1}$ & $\Delta f_{Random-1}$ & & No Context & $\Delta f_{Prev-1}$ & $\Delta f_{Next-1}$ & $\Delta f_{Random-1}$ & & No Context & $\Delta f_{Prev-1}$ & $\Delta f_{Next-1}$ & $\Delta f_{Random-1}$ \\
          \cmidrule{1-15}
          \multicolumn{15}{c}{\textbf{Function Words}} \\
          \cmidrule{1-15}
          PRON & 0.82 & -13.4\% & -13.4\% & -8.5\% & & 0.81 & -12.3\% & -13.6\% & -7.4\% & & 0.87 & -11.5\% & -11.5\% & -11.5\% \\
          PART & 0.66 & -7.6\% & -4.5\% & -6.1\% & & 0.66 & -10.6\% & -6.1\% & -6.1\% & & 0.76 & -11.8\% & -10.5\% & -11.8\% \\
          AUX & 0.87 & -6.9\% & -8.0\% & -5.7\% & & 0.67 & -9.0\% & -7.5\% & -6.0\% & & 0.9 & -5.6\% & -5.6\% & -3.3\% \\
          SCONJ & 0.9 & -5.6\% & -5.6\% & -5.6\% & & 0.84 & -9.5\% & -8.3\% & -7.1\% & & 0.93 & -6.5\% & -6.5\% & -6.5\% \\
          DET & 0.88 & -5.7\% & -4.5\% & -2.3\% & & 0.74 & -6.8\% & -6.8\% & -4.1\% & & 0.57 & -5.3\% & -7.0\% & -5.3\% \\
          ADP & 0.83 & -4.8\% & -3.6\% & -3.6\% & & 0.68 & -4.4\% & -5.9\% & -4.4\% & & 0.72 & -8.3\% & -8.3\% & -8.3\% \\
          CCONJ & 0.94 & -3.2\% & -4.3\% & -2.1\% & & 0.78 & -6.4\% & -6.4\% & -5.1\% & & 0.97 & -1.0\% & -1.0\% & 0.0\% \\
          \midrule
          \textbf{Average} & \textbf{0.84} & \textbf{-6.7\%} &\textbf{ -6.3\%} & \textbf{-4.8\%} & & \textbf{0.74} &\textbf{ -8.4\%} & \textbf{-7.8\%} & \textbf{-5.7\%} & & \textbf{0.82} & \textbf{-7.1\%} & \textbf{-7.2\%} & \textbf{-6.7\%} \\
          \midrule
          \multicolumn{15}{c}{\textbf{Content Words}} \\
          \cmidrule{1-15}
          ADV & 0.81 & -7.4\% & -7.4\% & -4.9\% & & 0.76 & -6.6\% & -6.6\% & -3.9\% & & 0.91 & -8.8\% & -5.5\% & -6.6\% \\
          VERB & 1.01 & -5.9\% & -5.9\% & -4.0\% & & 0.87 & -9.2\% & -8.0\% & -5.7\% & & 1.02 & -8.8\% & -8.8\% & -7.8\% \\
          NOUN & 1.09 & -4.6\% & -3.7\% & -2.8\% & & 0.89 & -5.6\% & -4.5\% & -2.2\% & & 0.99 & -7.1\% & -7.1\% & -4.0\% \\
          ADJ & 1.01 & -4.0\% & -3.0\% & -1.0\% & & 0.92 & -4.3\% & -4.3\% & -3.3\% & & 1.02 & -6.9\% & -6.9\% & -4.9\% \\
          PROPN & 1.02 & -2.0\% & -2.0\% & -2.0\% & & 0.95 & 0.0\% & 0.0\% & 0.0\% & & 1.02 & -1.0\% & -2.9\% & +1.0\% \\
          \midrule
          \textbf{Average} & \textbf{0.99} &\textbf{ -4.8\%} & \textbf{-4.4\%} & \textbf{-2.9\%} & & \textbf{0.88} & \textbf{-5.1\%} & \textbf{-4.7\%} & \textbf{-3.0\%} & & \textbf{0.99} & \textbf{-6.5\%} & \textbf{-6.2\%} & \textbf{-4.5\%} \\
         \bottomrule
    \end{tabular}
    }
    \caption{PoS-tag-level average fertility values for De $\rightarrow$ En, En $\rightarrow$, and En $\rightarrow$ Hi language pairs for different context settings. Values in the \texttt{No Context} column are the raw average fertility values and the values in $\Delta f_{Prev-1}$, $\Delta f_{Next-1}$, and $\Delta f_{Random-1}$ columns shows the relative difference between \texttt{No Context} and the corresponding context setting values in terms of percentage. $-$ and $+$ signs denote decrement and increment in average fertility values, respectively.}
    \label{tab:source_relative_fertility_scores}
\end{table*}

We analyse the fertility distribution across different context settings. As mentioned in Section \ref{sec:fertility_computation}, we use \texttt{Stanza} to tag source data with universal PoS tags\footnote{Universal PoS-tag list is presented in Table \ref{tab:upos} in the Appendix.}. Specifically, we calculate the relative difference between \texttt{No-Context} ($f_{No-Context}$) setting fertility and a specific context ($f_{context}$) setting fertility as:

\begin{equation}
    \Delta f_{context}\,=\,\frac{f_{context}\,-\,f_{No-Context}}{f_{No-Context}}\,*\,100
\end{equation}

\begin{figure*}[!ht]
    \centering
    \subfigure[]{\includegraphics[width=0.32\textwidth]{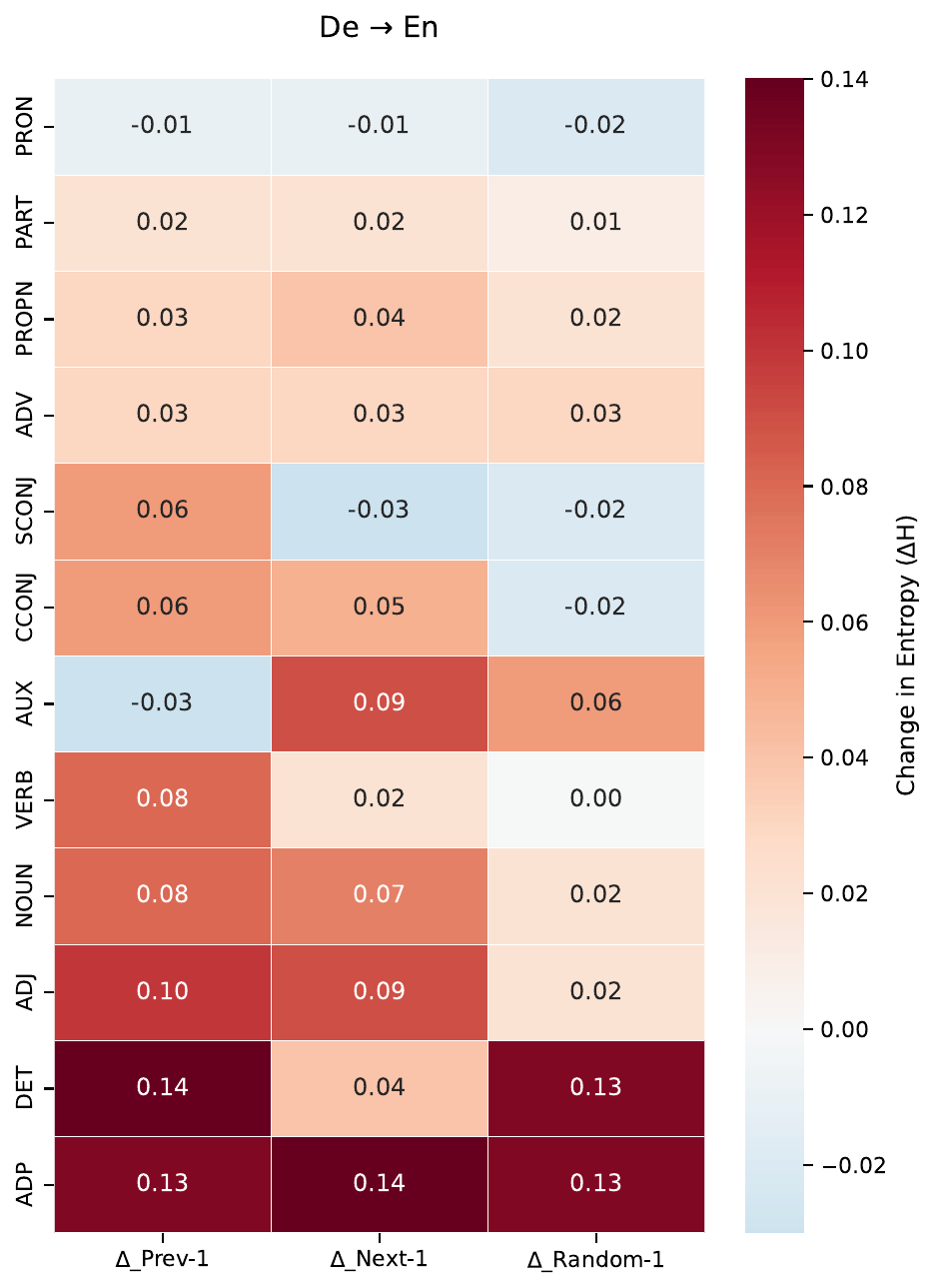}}
    \subfigure[]{\includegraphics[width=0.32\textwidth]{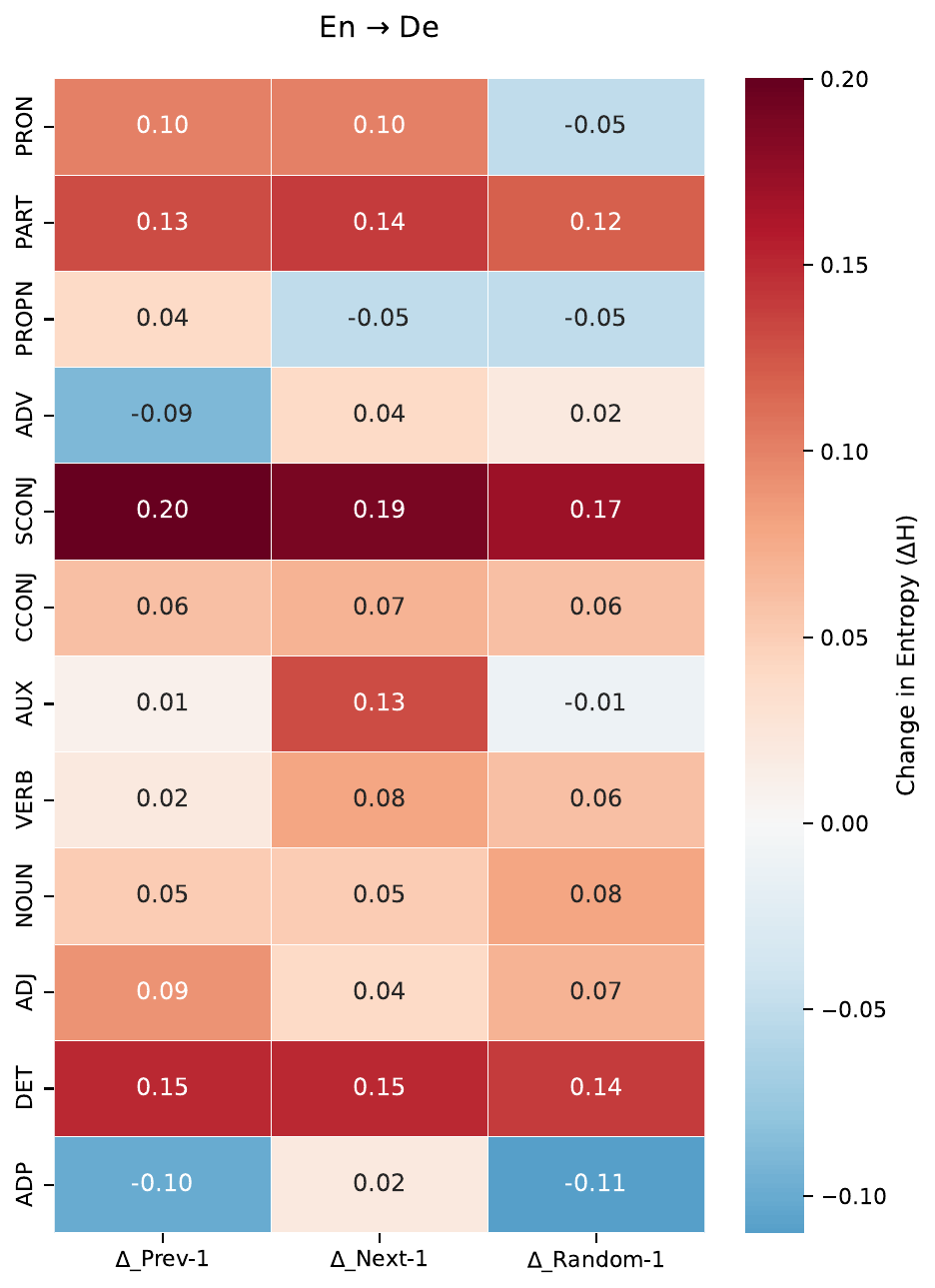}}
    \subfigure[]{\includegraphics[width=0.32\textwidth]{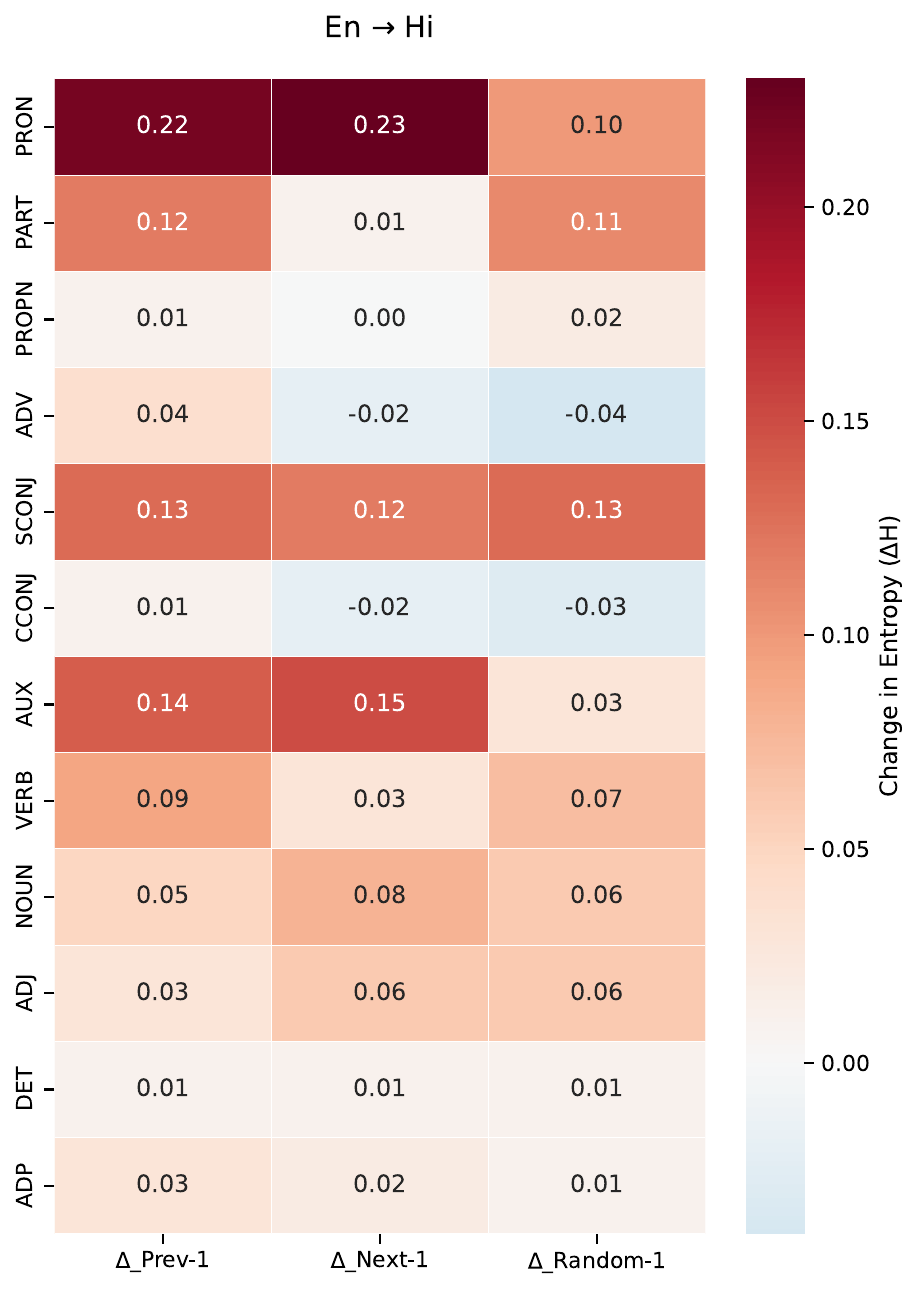}}
    \caption{Entropy heatmaps of the PoS categories for different context settings. Figures (a), (b), and (c) represent the relative difference for De $\rightarrow$ En, En $\rightarrow$ De, and En $\rightarrow$ Hi language pairs, respectively. The Y-axis shows various PoS categories, and the X-axis shows different context settings. Negative (represented in blue colour) and positive (represented in red colour) values show a decrease and an increase in entropy compared to the \texttt{No-Context} setting.}
    \label{fig:source_entropy_heatmaps}
\end{figure*}

Table \ref{tab:source_relative_fertility_scores} shows the relative percentage difference\footnote{Raw average fertility values for different context settings are presented in Table \ref{tab:source_raw_fertility_scores} in the Appendix.}. Negative sign indicates a decrease in average fertility value, suggesting that some generation responsibility is shifted away from source tokens when context is introduced. To better understand how this redistribution occurs across linguistic categories, we group PoS tags into function words and content words. We observe that function words exhibit larger fertility reductions compared to content words when context is added. This suggests that context primarily affects grammatical and discourse-related tokens. Among specific PoS categories, pronouns (PRON), particles (PART), subordinating conjunctions (SCONJ), adverbs (ADV), and verbs (VERB) exhibit the largest fertility reductions. This observation is in line with previous studies showing that contextual information is particularly important for translating discourse-level phenomena \citep{voita-etal-2018-context,kim-etal-2019-document,fernandes-etal-2023-translation,castilho2025survey}.

In contrast, fertility values for proper nouns (PROPN) remain nearly identical to the \texttt{No-Context} setting, indicating that context has minimal effect on named entity translation. For the random context setting ($\Delta f_{Random-1}$), we still observe reductions in source fertility. However, the magnitude is smaller than when the context is either the previous or the next sentence. This behaviour is consistent with the corpus coverage analysis (cf. Figure \ref{fig:sentence_coverage}), where both source and target coverage increase even when the context is randomly sampled. We next examine the stability of these PoS-level fertility patterns through entropy analysis.

\subsection{Source PoS Entropy Analysis}
\label{sec:source_pos_entropy_analysis}
While PoS-tag-level fertility values reveal how translation responsibility is redistributed across tokens, it is also important to examine how stable these patterns remain when context is introduced. To study this, we compute the entropy of the fertility distributions for each PoS category as described in Section \ref{sec:entropy_of_fertility} (cf. Equation \ref{eq:pos_entropy}). Specifically, we measure the difference between the entropy values of the \texttt{No-Context} and a given context setting as:

\begin{equation}
    \Delta \mathcal{H}\,=\,\mathcal{H}_{context}\,-\,\mathcal{H}_{No-Context}
\end{equation}

Figure \ref{fig:source_entropy_heatmaps} presents heatmaps of these entropy differences across PoS categories\footnote{Raw entropy values for different context settings are presented in Table \ref{tab:source_raw_entropy_scores} in the Appendix.}. An increase in entropy indicates that the fertility distribution for a given PoS category becomes more variable (cf. Equation \ref{eq:pos_entropy}) when context is added, whereas a decrease suggests more stable fertility behaviour.

From the entropy heatmaps, we observe that proper nouns (PROPN), adverbs (ADV), and nouns (NOUN) remain largely stable across context settings. However, the effect of context varies across language pairs. For the De $\rightarrow$ En language pair, context has little effect on pronouns (PRON) and particles (PART), while determiners (DET) and adpositions (ADP) show larger entropy increases. In contrast, for the En $\rightarrow$ De and En $\rightarrow$ Hi language pairs, entropy changes are relatively smaller for adpositions, while pronouns and subordinating conjunctions (SCONJ) exhibit slightly larger variability. These differences across language pairs likely reflect structural and typological differences between the languages, such as variation in morphological richness, word order flexibility, and the way discourse relations are expressed.

Overall, function words exhibit higher average entropy changes (+0.07) compared to content words (+0.04) across all language and context settings. This observation suggests that context tends to influence the behaviour of grammatical and discourse-related tokens more than content tokens. We also observe that the average entropy difference between the \texttt{Random-1} and \texttt{No-Context} settings (+0.04) is slightly lower than the difference observed for the \texttt{Prev-1} and \texttt{Next-1} settings (+0.06). This indicates that the randomly sampled context introduces comparatively less variability in fertility behaviour. To further investigate how translation responsibility shifts from source to context tokens, we analyse the amount of zero-fertility tokens.

\begin{figure*}[!ht]
    \centering
    \subfigure[]{\includegraphics[width=0.32\textwidth]{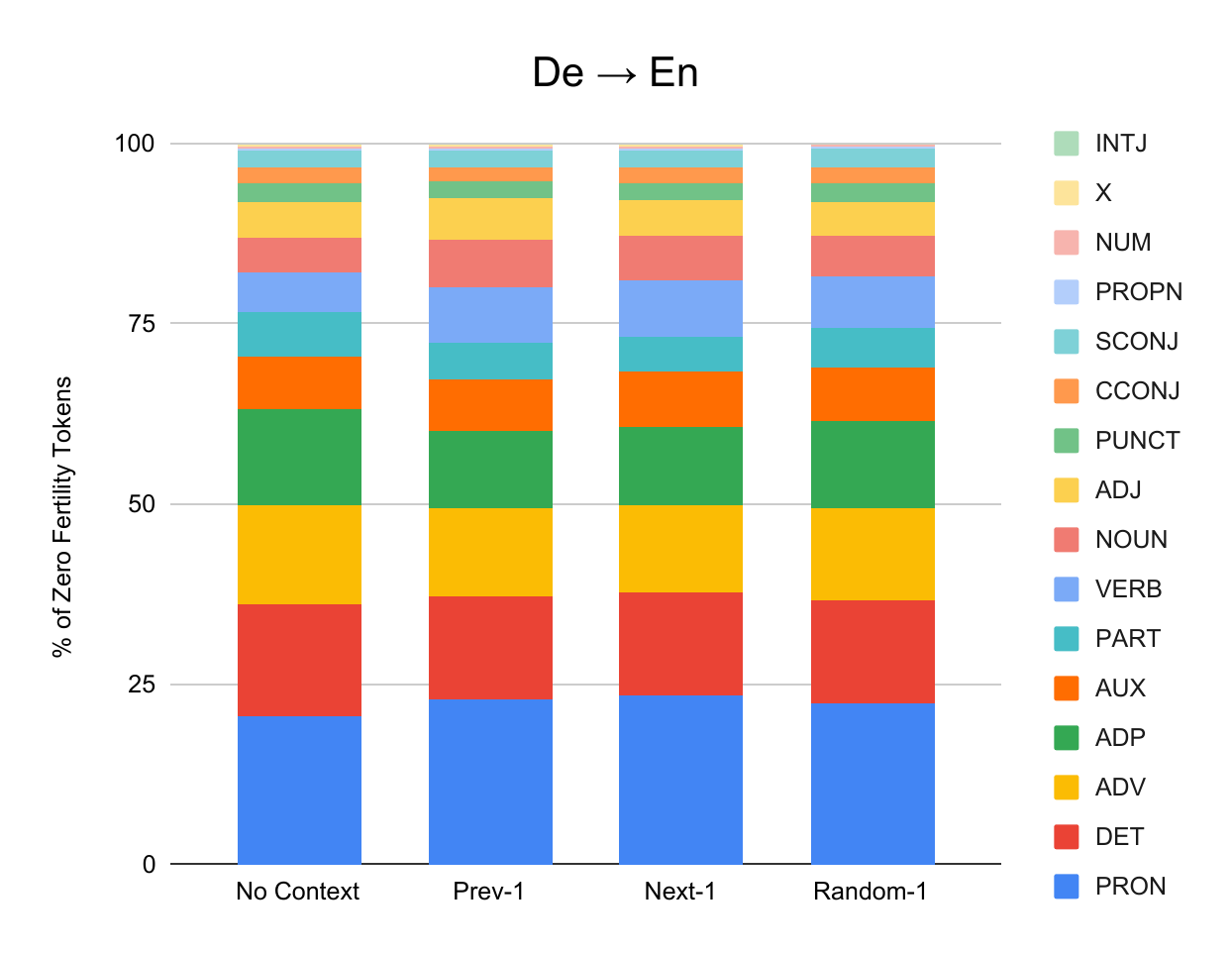}}
    \subfigure[]{\includegraphics[width=0.32\textwidth]{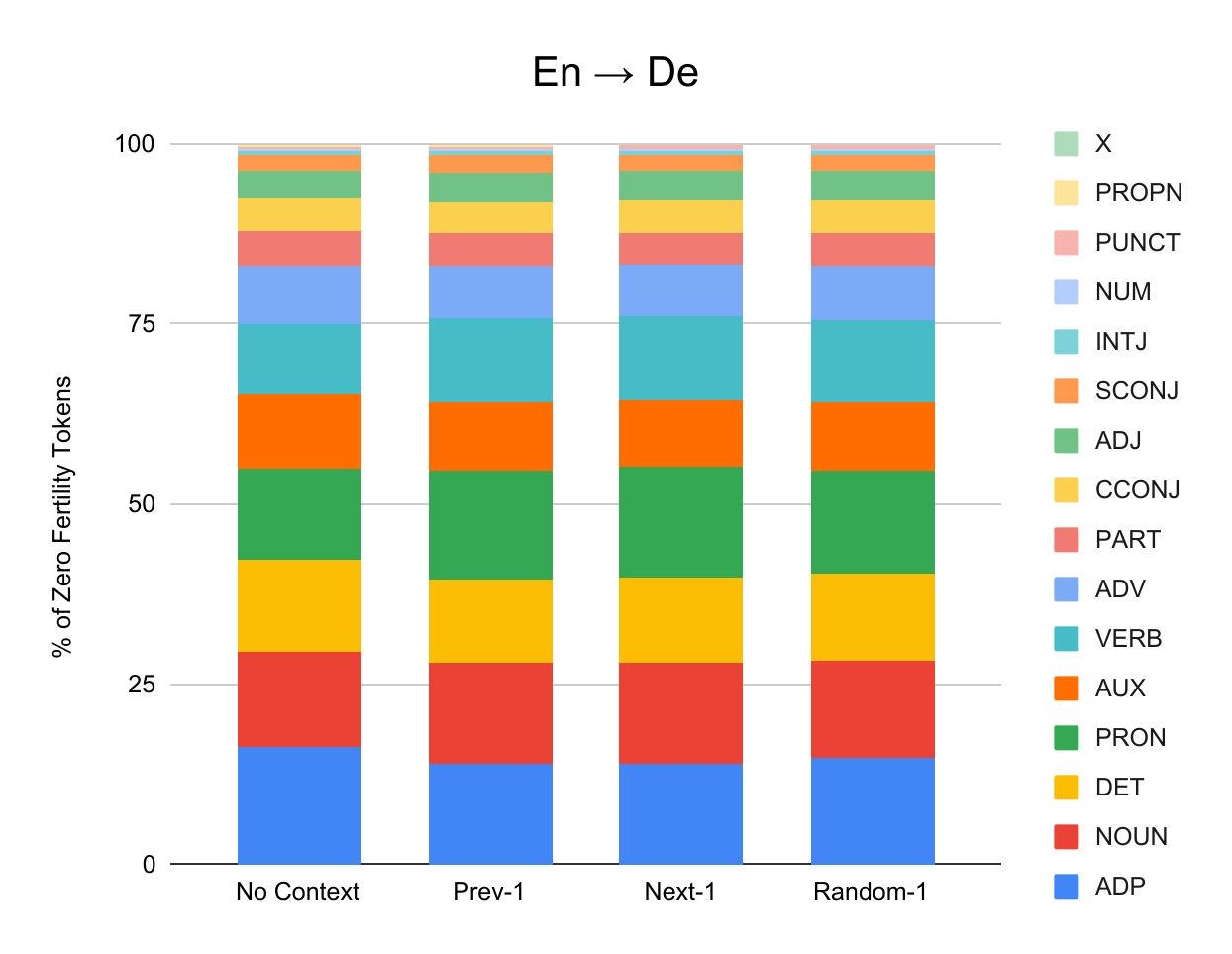}}
    \subfigure[]{\includegraphics[width=0.32\textwidth]{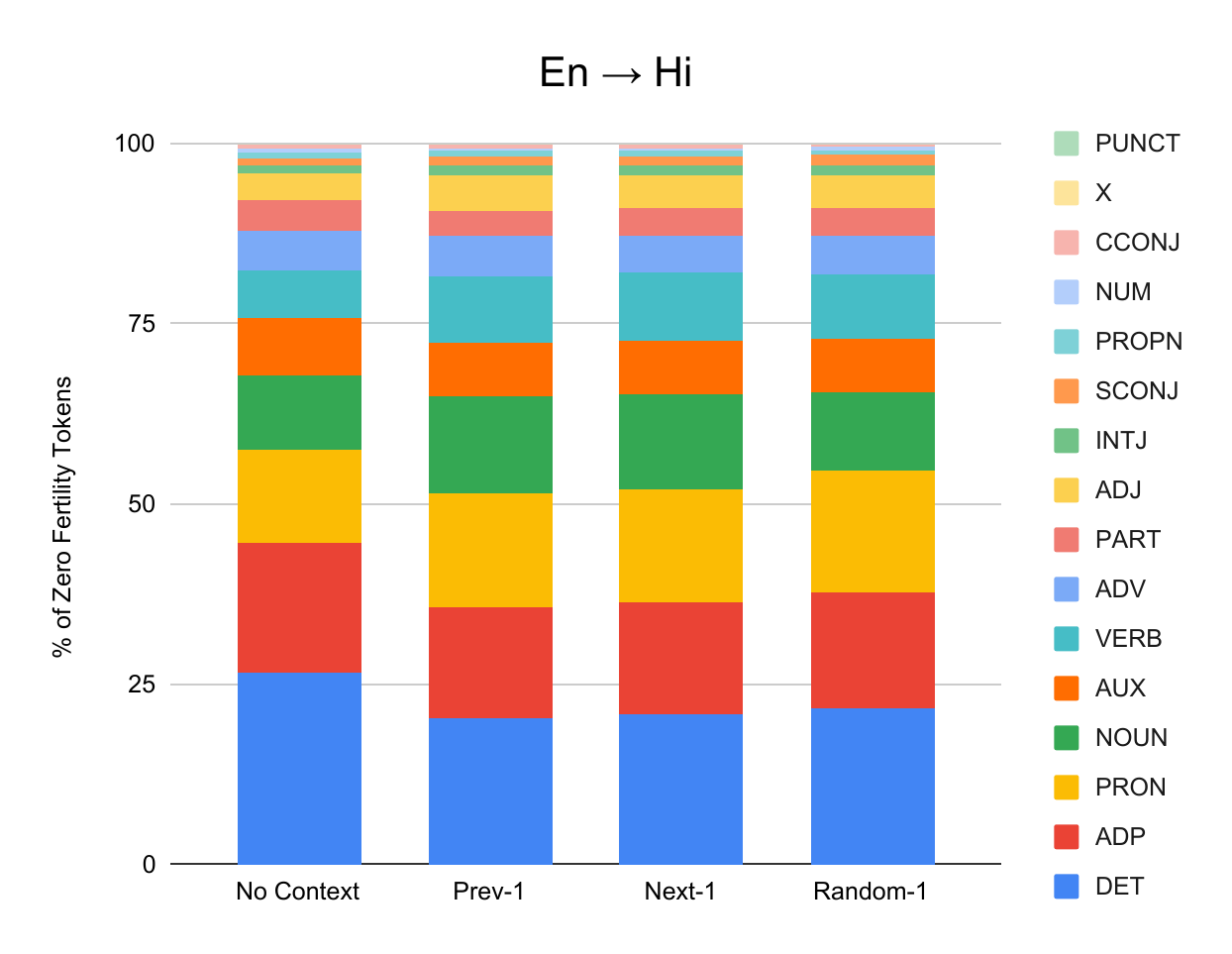}}
    \caption{Percentage of zero fertility tokens per PoS category that contribute to total zero fertility tokens for different context settings. Figures (a), (b), and (c) represent statistics for De $\rightarrow$ En, En $\rightarrow$ De, and En $\rightarrow$ Hi language pairs, respectively.}
    \label{fig:source_zero_fertility_coverage}
\end{figure*}

\subsection{Zero-Fertility Coverage Analysis}
\label{sec:source_zero_fertility_analysis}
We hypothesise that introducing context will increase the number of source tokens that lose direct alignment links, as some of the generation responsibility will be shifted to context tokens. To examine this behaviour, we measure the percentage of tokens that produce no alignment links (zero-fertility tokens) across different context settings.

\begin{table}[!ht]
    \centering
    \resizebox{1.0\linewidth}{!}{
    \begin{tabular}{ccccc}
        \toprule
        \textbf{Lang Pair }& \textbf{No Context} & \textbf{Prev-1} & \textbf{Next-1} & \textbf{Random-1} \\
        \midrule
        De $\rightarrow$ En & 11.7 & 16.8 & 16.7 & 15.0 \\
        \cmidrule{1-5}
        En $\rightarrow$ De & 19.8 & 25.7 & 25.4 & 23.5 \\
        \cmidrule{1-5}
        En $\rightarrow$ Hi & 15.5 & 21.8 & 21.7 & 20.5 \\
        \bottomrule
    \end{tabular}
    }
    \caption{Percentage of zero fertility source tokens for different context settings.}
    \label{tab:source_zero_fertility_percentage}
\end{table}

Table \ref{tab:source_zero_fertility_percentage} presents the corpus-level percentages of zero-fertility tokens. From the results, we observe that adding context increases the proportion of zero-fertility tokens by approximately 3.3–6.3\% across different context settings. This increase suggests that the addition of context reduces the direct contribution of some source tokens to target generation, which is consistent with the notion of translation responsibility being redistributed across the context-added input sequence.

We further analyse the PoS-level behaviour by examining the contribution of different PoS categories to the overall pool of zero-fertility tokens. Figure \ref{fig:source_zero_fertility_coverage} shows the distribution of zero-fertility tokens across PoS categories\footnote{Raw zero-fertility values for different context settings are presented in Table \ref{tab:pos_zero_fertility_raw_scores} in the Appendix.}. In general, pronouns (PRON), determiners (DET), adpositions (ADP), and auxiliary verbs (AUX) contribute a larger proportion of zero-fertility tokens. This observation aligns with our earlier fertility analysis (cf. Table \ref{tab:source_relative_fertility_scores}), which showed larger reductions in average fertility values for these categories. We next analyse the average fertility and entropy patterns for context tokens, as the generation responsibility partially shifts toward them.

\subsection{Context Fertility and Entropy Analysis}
We further examine how context tokens participate in target generation by analysing their average fertility and entropy across different context settings. Since only a small proportion of context tokens form alignment links with target tokens (cf. Figure \ref{fig:sentence_coverage}), we compute these statistics using only context tokens with at least one alignment link (fertility $> 0$). This allows us to study the behaviour of context tokens that actively contribute to translation.

\begin{table}[!ht]
    \centering
    \resizebox{1.0\linewidth}{!}{
    \begin{tabular}{ccccccc}
        \toprule
        \multirow{2}*{\textbf{Lang Pair}} & \multicolumn{2}{c}{\textbf{Prev-1}} & \multicolumn{2}{c}{\textbf{Next-1}} & \multicolumn{2}{c}{\textbf{Random-1}} \\
        \cmidrule{2-7}
         & \textbf{Fertility} & \textbf{Entropy} & \textbf{Fertility} & \textbf{Entropy} & \textbf{Fertility} & \textbf{Entropy} \\
        \midrule
        De $\rightarrow$ En & 1.03 & 0.15 & 1.04 & 0.18 & 1.05 & 0.26 \\
        \cmidrule{1-7}
        En $\rightarrow$ De & 1.02 & 0.09 & 1.02 & 0.14 & 1.03 & 0.15 \\
        \cmidrule{1-7}
        En $\rightarrow$ Hi & 1.12 & 0.24 & 1.09 & 0.24 & 1.10 & 0.19 \\
        \bottomrule
    \end{tabular}
    }
    \caption{Average corpus-level fertility and entropy values for context tokens with different context settings.}
    \label{tab:context_avg_fertility_entropy}
\end{table}

\begin{figure*}[!ht]
    \centering
    \subfigure[]{\includegraphics[width=0.45\textwidth]{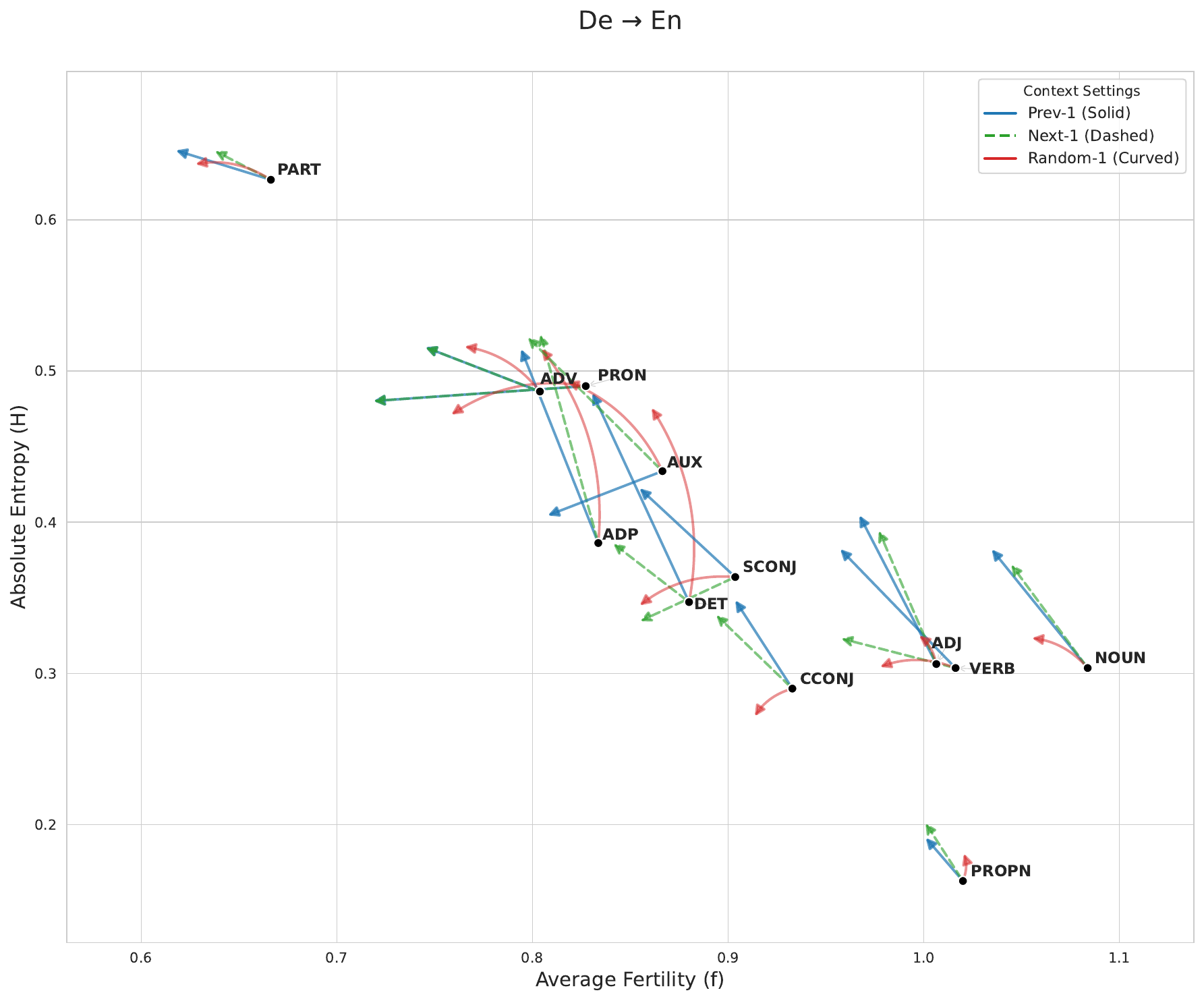}}
    \subfigure[]{\includegraphics[width=0.45\textwidth]{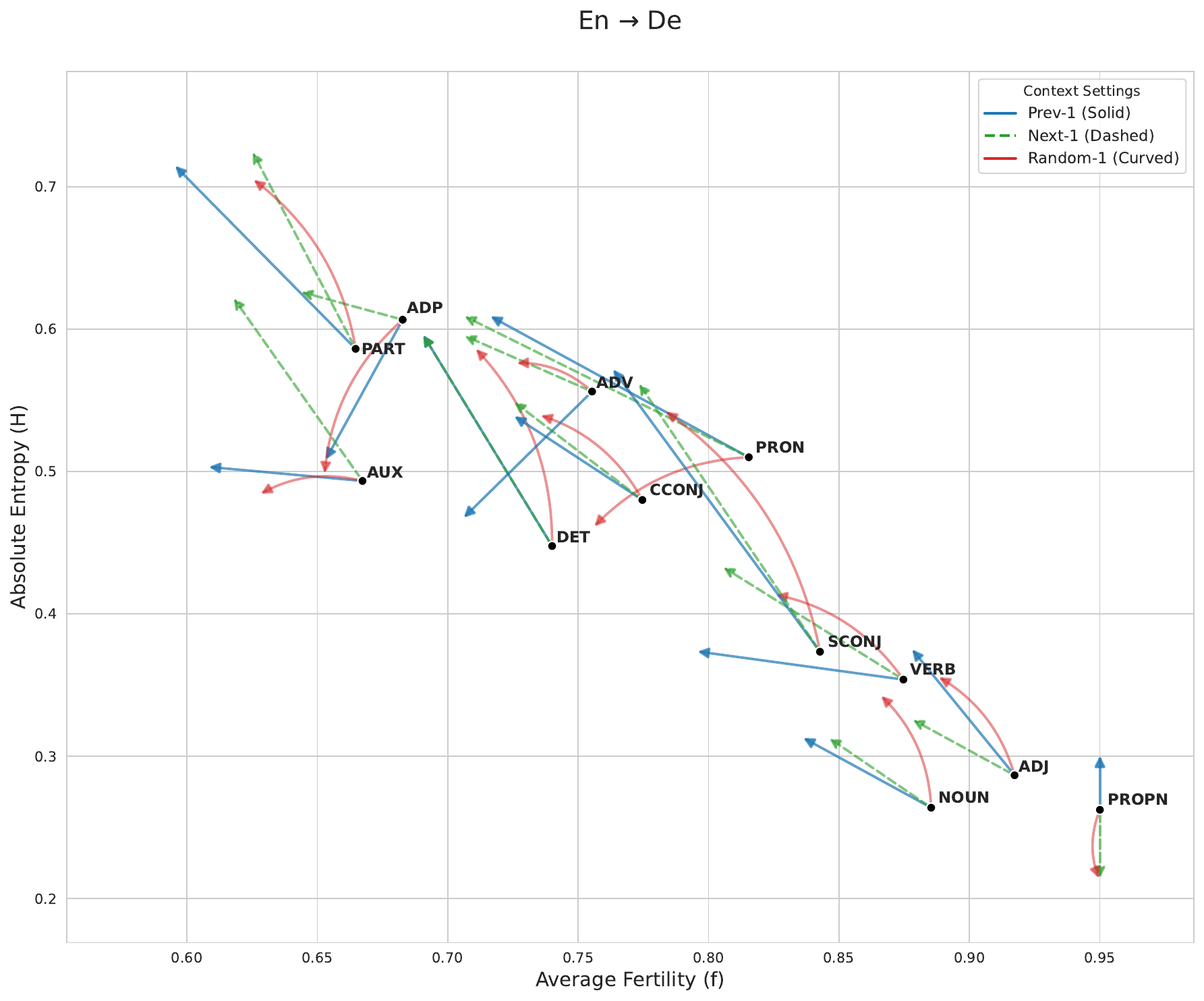}}
    \subfigure[]{\includegraphics[width=0.45\textwidth]{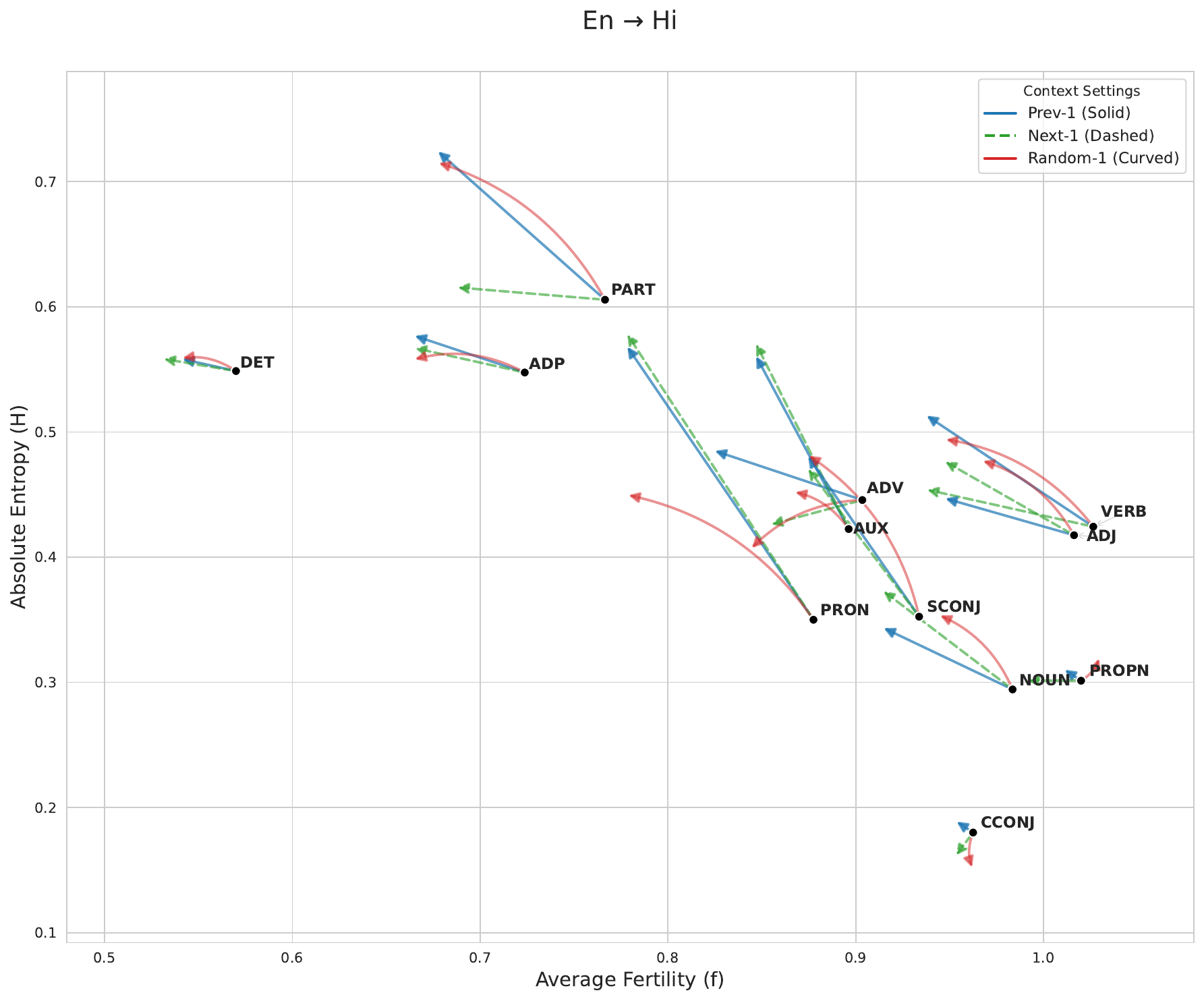}}
    \caption{Joint fertility-entropy plots of various PoS categories across context settings. Figures (a), (b), and (c) represent joint plots for De $\rightarrow$ En, En $\rightarrow$ De, and En $\rightarrow$ Hi language pairs, respectively. The Y-axis shows absolute entropy, and the X-axis shows average fertility values. Values for \textbf{No-Context} setting are shown as points, and the arrows originating from these points represent variation for different context settings.}
    \label{fig:source_fertility_entropy_quiver_plots}
\end{figure*}

Table \ref{tab:context_avg_fertility_entropy} shows the corpus-level average fertility and entropy values for context tokens\footnote{Raw values for different context settings are presented in Tables \ref{tab:context_raw_fertility_scores} and \ref{tab:context_raw_entropy_scores} in the Appendix.} across language pairs and context settings. We observe that the average context token fertility remains close to 1, with relatively low entropy. This suggests that when context tokens participate in translation, they typically generate approximately one target token. The low entropy values further indicate that the fertility distributions of context tokens remain relatively stable across sentences. Together, these observations suggest that when translation responsibility shifts from source tokens to context tokens, the latter tend to contribute in a consistent manner rather than generating multiple target tokens. This behaviour also explains why the overall corpus-level fertility remains largely unchanged (cf. Table \ref{tab:avg_corpus_fertility}) even when context is introduced.

\subsection{Unifying Fertility and Entropy Measures}
In the previous sections, we analysed the behaviour of source PoS categories with respect to fertility (cf. Section \ref{sec:source_pos_fertility_analysis}) and entropy (cf. Section \ref{sec:source_pos_entropy_analysis}) independently. Here, we combine these two measures to examine how the structural contribution of tokens (fertility) relates to the stability of their translation behaviour (entropy). Figure \ref{fig:source_fertility_entropy_quiver_plots} presents the joint fertility–entropy plots for different context settings. The \texttt{No-Context} values for each PoS category are shown as points, while the changes introduced by various context settings are denoted using arrows originating from these points. The direction and length of the arrows represent the direction of change and magnitude in fertility and entropy when context is introduced.

Across most PoS categories and language pairs, we observe a consistent pattern where the addition of context results in a reduction in average fertility with an increase in entropy. This behaviour is consistent with our earlier findings that adding context redistributes a portion of the generative responsibility away from source tokens. The joint plots also reveal a clear relationship between fertility and entropy stability. PoS categories with higher average fertility values tend to exhibit smaller entropy variations, indicating more stable translation behaviour. In contrast, categories with lower fertility values show larger entropy shifts when context is introduced. This observation aligns with our earlier analysis where content words (e.g., nouns and proper nouns) remain relatively stable, while function words (e.g., pronouns, determiners, and adpositions) exhibit larger variability. Overall, this joint analysis highlights that context primarily affects tokens with lower structural contribution, while tokens with higher fertility remain largely stable.

\section{Conclusion}
In this work, we proposed a post-hoc framework for quantifying selective context usage in translation using fertility and entropy measures derived from word alignments. Our central finding is that context does not simply add new information; rather, it redistributes generative responsibility from source tokens to context tokens. This redistribution is selective: function words, such as pronouns, auxiliary verbs, particles, and conjunctions, exhibit the largest reductions in fertility when context is introduced, while high-fertility content words (nouns, proper nouns) remain stable. This pattern is consistent across three typologically distinct language pairs (German $\leftrightarrow$ English, English $\rightarrow$ Hindi) and four context settings (no context, previous sentence, next sentence, random sentence).

The entropy analysis further reinforces this selective-use interpretation. Context-sensitive tokens show higher entropy (greater variability), while stable tokens show low entropy, suggesting that context resolves ambiguity rather than adding new lexical information. Moreover, when tokens from the context participate in translation, they consistently generate approximately one target token with low entropy, indicating focused, one-to-one selective attention. These patterns collectively point to a model of translation where context is invoked sparingly and precisely, rather than densely and indiscriminately.

Our framework offers several advantages over existing approaches. It is model-agnostic (requiring only alignments and PoS tags), discourse-agnostic (not limited to predefined phenomena), and lightweight (no model internals or expensive inference required). It provides a diagnostic baseline for understanding which tokens are genuinely context-sensitive in human translation, a ground truth against which machine translation models can be evaluated.

In future work, we plan to apply our framework to machine translation outputs, comparing fertility-entropy patterns in model-generated translations to the human baselines established in our current work. We also aim to correlate our measures with model attention patterns, to test whether the tokens that models attend to correspond to the tokens that humans align with. Finally, we plan to extend our analysis to additional typologically diverse languages (e.g., agglutinative languages) and to investigate semantic and discourse-level phenomena beyond the lexical and syntactic levels studied here.

\section{Limitations}
Our current work has three limitations:

\paragraph{Dependency on existing toolkits:} The proposed approach relies on token-level alignments extracted using the \texttt{awesome-align} and PoS tagging obtained using the \texttt{Stanza} toolkits. Although we use the pre-trained models provided by these toolkits, the accuracy of our analysis depends on the quality of the generated alignments and PoS tags. Furthermore, zero-fertility source tokens do not necessarily imply the absence of context influence. In some cases, context dependency may be resolved implicitly through intra-sentential dependencies that are not captured by direct alignment links.

\paragraph{Language-pair diversity:} Our experiments are conducted on language pairs from both related (English–German) and unrelated (English–Hindi) language families. While the observed patterns remain consistent across these settings, additional experiments on typologically diverse languages (e.g., morphologically rich or agglutinative languages) could provide further insights into how context utilisation varies across linguistic structures.

\paragraph{Analysis limited to lexical and syntactic levels:} Our current work is limited to lexical and syntactic levels. Investigation of the same phenomenon at deeper semantic or discourse levels remains beyond the scope of the present study and is an important direction for future work.

\bibliography{custom}

\appendix
\section{Appendix}
\label{sec:appendix}

\begin{table}[!ht]
    \centering
    \resizebox{0.75\linewidth}{!}{
    \begin{tabular}{c|l}
        \toprule
        \textbf{Tag} & \textbf{Description} \\
        \midrule
        ADJ & Adjective \\
        ADP & Adposition \\
        ADV & Adverb \\
        AUX & Auxiliary Verb \\
        CCONJ & Coordinating Conjunction \\
        DET & Determiner \\
        INTJ & Interjection \\
        NOUN & Noun \\
        NUM & Numeral \\
        PART & Particle \\
        PRON & Pronoun \\
        PROPN & Proper Noun \\
        PUNCT & Punctuation \\
        SCONJ & Subordinating Conjunction \\
        SYM & Symbol \\
        VERB & Verb \\
        X & Other \\
        \bottomrule
    \end{tabular}
    }
    \caption{Universal Part-of-Speech Categories}
    \label{tab:upos}
\end{table}


\begin{table*}[!ht]
    \centering
    \resizebox{1.0\linewidth}{!}{
    \begin{tabular}{l||ccccc||ccccc||cccc}
         \toprule
         \multirow{2}*{\textbf{PoS Tag}} & \multicolumn{4}{c}{\textbf{De $\rightarrow$ En}} & & \multicolumn{4}{c}{\textbf{En $\rightarrow$ De}} & & \multicolumn{4}{c}{\textbf{En $\rightarrow$ Hi}} \\
         \cmidrule{2-15}
          & No Context & Prev-1 & Next-1 & Random-1 & & No Context & Prev-1 & Next-1 & Random-1 & & No Context & Prev-1 & Next-1 & Random-1 \\
          \cmidrule{1-15}
          PRON & 0.82 & 0.71 & 0.71 & 0.75 &  & 0.81 & 0.71 & 0.70 & 0.75 &  & 0.87 & 0.77 & 0.77 & 0.77 \\
          VERB & 1.01 & 0.95 & 0.95 & 0.97 &  & 0.87 & 0.79 & 0.80 & 0.82 &  & 1.02 & 0.93 & 0.93 & 0.94 \\
          ADP & 0.83 & 0.79 & 0.80 & 0.80 &  & 0.68 & 0.65 & 0.64 & 0.65 &  & 0.72 & 0.66 & 0.66 & 0.66 \\
          DET & 0.88 & 0.83 & 0.84 & 0.86 &  & 0.74 & 0.69 & 0.69 & 0.71 &  & 0.57 & 0.54 & 0.53 & 0.54 \\
          ADJ & 1.01 & 0.97 & 0.98 & 1.00 &  & 0.92 & 0.88 & 0.88 & 0.89 &  & 1.02 & 0.95 & 0.95 & 0.97 \\
          NOUN  & 1.09 & 1.04 & 1.05 & 1.06 &  & 0.89 & 0.84 & 0.85 & 0.87 &  & 0.99 & 0.92 & 0.92 & 0.95 \\
          CCONJ & 0.94 & 0.91 & 0.90 & 0.92 &  & 0.78 & 0.73 & 0.73 & 0.74 &  & 0.97 & 0.96 & 0.96 & 0.97 \\
          ADV & 0.81 & 0.75 & 0.75 & 0.77 &  & 0.76 & 0.71 & 0.71 & 0.73 &  & 0.91 & 0.83 & 0.86 & 0.85 \\
          AUX & 0.87 & 0.81 & 0.80 & 0.82 &  & 0.67 & 0.61 & 0.62 & 0.63 &  & 0.90 & 0.85 & 0.85 & 0.87 \\
          PROPN & 1.02 & 1.00 & 1.00 & 1.02 &  & 0.95 & 0.95 & 0.95 & 0.95 &  & 1.02 & 1.01 & 0.99 & 1.03 \\
          SCONJ & 0.90 & 0.85 & 0.85 & 0.85 &  & 0.84 & 0.76 & 0.77 & 0.78 &  & 0.93 & 0.87 & 0.87 & 0.87 \\
          PART & 0.66 & 0.61 & 0.63 & 0.62 &  & 0.66 & 0.59 & 0.62 & 0.62 &  & 0.76 & 0.67 & 0.68 & 0.67 \\
          NUM & 0.98 & 0.96 & 0.96 & 0.98 &  & 0.97 & 0.94 & 0.96 & 0.97 &  & 0.95 & 0.95 & 0.94 & 0.95 \\
          PUNCT & 0.85 & 0.81 & 0.82 & 0.83 &  & 0.90 & 0.86 & 0.85 & 0.85 &  & 2.50 & 2.50 & 2.38 & 2.25 \\
          INTJ & 1.03 & 0.88 & 1.03 & 0.97 &  & 0.78 & 0.74 & 0.71 & 0.78 &  & 0.95 & 0.86 & 0.86 & 0.85 \\
          X & 0.92 & 0.94 & 0.90 & 0.95 &  & 0.94 & 0.87 & 0.98 & 1.02 &  & 1.36 & 1.50 & 1.21 & 1.36 \\
          SYM & - & - & - & - &  & 1.00 & 1.00 & 0.93 & 1.00 &  & 1.43 & 1.43 & 1.43 & 1.43 \\
         \bottomrule
    \end{tabular}
    }
    \caption{Source PoS-tag-level raw fertility values for De $\rightarrow$ En, En $\rightarrow$ De, and En $\rightarrow$ Hi language pairs.}
    \label{tab:source_raw_fertility_scores}
\end{table*}


\begin{table*}[!ht]
    \centering
    \resizebox{1.0\linewidth}{!}{
    \begin{tabular}{l||ccccc||ccccc||cccc}
         \toprule
         \multirow{2}*{\textbf{PoS Tag}} & \multicolumn{4}{c}{\textbf{De $\rightarrow$ En}} & & \multicolumn{4}{c}{\textbf{En $\rightarrow$ De}} & & \multicolumn{4}{c}{\textbf{En $\rightarrow$ Hi}} \\
         \cmidrule{2-15}
          & No Context & Prev-1 & Next-1 & Random-1 & & No Context & Prev-1 & Next-1 & Random-1 & & No Context & Prev-1 & Next-1 & Random-1 \\
          \cmidrule{1-15}
          PRON & 0.49 & 0.48 & 0.48 & 0.47 &  & 0.51 & 0.61 & 0.61 & 0.46 &  & 0.35 & 0.57 & 0.58 & 0.45 \\
          VERB & 0.30 & 0.38 & 0.32 & 0.30 &  & 0.35 & 0.37 & 0.43 & 0.41 &  & 0.42 & 0.51 & 0.45 & 0.49 \\
          ADP & 0.38 & 0.51 & 0.52 & 0.51 &  & 0.60 & 0.50 & 0.62 & 0.49 &  & 0.54 & 0.57 & 0.56 & 0.55 \\
          DET & 0.34 & 0.48 & 0.38 & 0.47 &  & 0.44 & 0.59 & 0.59 & 0.58 &  & 0.54 & 0.55 & 0.55 & 0.55 \\
          ADJ & 0.30 & 0.40 & 0.39 & 0.32 &  & 0.28 & 0.37 & 0.32 & 0.35 &  & 0.41 & 0.44 & 0.47 & 0.47 \\
          NOUN  & 0.30 & 0.38 & 0.37 & 0.32 &  & 0.26 & 0.31 & 0.31 & 0.34 &  & 0.29 & 0.34 & 0.37 & 0.35 \\
          CCONJ & 0.29 & 0.35 & 0.34 & 0.27 &  & 0.48 & 0.54 & 0.55 & 0.54 &  & 0.18 & 0.19 & 0.16 & 0.15 \\
          ADV & 0.49 & 0.52 & 0.52 & 0.52 &  & 0.56 & 0.47 & 0.60 & 0.58 &  & 0.45 & 0.49 & 0.43 & 0.41 \\
          AUX & 0.44 & 0.41 & 0.53 & 0.50 &  & 0.50 & 0.51 & 0.63 & 0.49 &  & 0.43 & 0.57 & 0.58 & 0.46 \\
          PROPN & 0.17 & 0.20 & 0.21 & 0.19 &  & 0.27 & 0.31 & 0.22 & 0.22 &  & 0.31 & 0.32 & 0.31 & 0.33 \\
          SCONJ & 0.37 & 0.43 & 0.34 & 0.35 &  & 0.38 & 0.58 & 0.57 & 0.55 &  & 0.36 & 0.49 & 0.48 & 0.49 \\
          PART & 0.63 & 0.65 & 0.65 & 0.64 &  & 0.59 & 0.72 & 0.73 & 0.71 &  & 0.61 & 0.73 & 0.62 & 0.72 \\
          NUM & 0.19 & 0.25 & 0.24 & 0.23 &  & 0.30 & 0.37 & 0.34 & 0.31 &  & 0.30 & 0.42 & 0.38 & 0.39 \\
          PUNCT & 0.43 & 0.57 & 0.47 & 0.48 &  & 0.42 & 0.39 & 0.54 & 0.54 &  & 0.91 & 0.91 & 0.95 & 0.77 \\
          INTJ & 0.49 & 0.52 & 0.49 & 0.34 &  & 0.62 & 0.60 & 0.58 & 0.61 &  & 0.34 & 0.51 & 0.42 & 0.41 \\
          X & 0.44 & 0.43 & 0.57 & 0.43 &  & 0.33 & 0.52 & 0.46 & 0.46 &  & 0.76 & 0.71 & 0.79 & 0.76 \\
          SYM & - & - & - & - &  & 0.99 & 0.99 & 0.99 & 0.99 &  & 0.99 & 0.99 & 0.99 & 0.99 \\
         \bottomrule
    \end{tabular}
    }
    \caption{Source PoS-tag-level raw entropy values for De $\rightarrow$ En, En $\rightarrow$ De, and En $\rightarrow$ Hi language pairs.}
    \label{tab:source_raw_entropy_scores}
\end{table*}


\begin{table*}[!ht]
    \centering
    \resizebox{1.0\linewidth}{!}{
    \begin{tabular}{l||cccc||cccc||ccc}
         \toprule
         \multirow{2}*{\textbf{PoS Tag}} & \multicolumn{3}{c}{\textbf{De $\rightarrow$ En}} & & \multicolumn{3}{c}{\textbf{En $\rightarrow$ De}} & & \multicolumn{3}{c}{\textbf{En $\rightarrow$ Hi}} \\
         \cmidrule{2-12}
          & Prev-1 & Next-1 & Random-1 & & Prev-1 & Next-1 & Random-1 & & Prev-1 & Next-1 & Random-1 \\
          \cmidrule{1-12}
          PRON & 1.02 & 1.01 & 1.02 &  & 1.02 & 1.01 & 1.02 &  & 1.01 & 1.03 & 1.02 \\
          VERB & 1.05 & 1.06 & 1.05 &  & 1.02 & 1.02 & 1.02 &  & 1.07 & 1.06 & 1.06 \\
          ADP & 1.01 & 1.03 & 1.02 &  & 1.01 & 1.02 & 1.01 &  & 1.03 & 1.04 & 1.03 \\
          DET & 1.03 & 1.02 & 1.02 &  & 1.01 & 1.01 & 1.01 &  & 1.03 & 1.02 & 1.04 \\
          ADJ & 1.05 & 1.05 & 1.06 &  & 1.02 & 1.04 & 1.06 &  & 1.07 & 1.08 & 1.03 \\
          NOUN  & 1.05 & 1.05 & 1.05 &  & 1.02 & 1.02 & 1.03 &  & 1.06 & 1.06 & 1.07 \\
          CCONJ & 1.02 & 1.01 & 1.01 &  & 1.01 & 1.02 & 1.00 &  & 1.00 & 1.04 & 1.00 \\
          ADV & 1.04 & 1.02 & 1.04 &  & 1.04 & 1.02 & 1.01 &  & 1.06 & 1.08 & 1.07 \\
          AUX & 1.01 & 1.02 & 1.03 &  & 1.02 & 1.02 & 1.02 &  & 1.05 & 1.03 & 1.04 \\
          PROPN & 1.03 & 1.06 & 1.02 &  & 1.00 & 1.00 & 1.14 &  & 1.09 & 1.05 & 1.09 \\
          SCONJ & 1.01 & 1.00 & 1.04 &  & 1.05 & 1.02 & 1.03 &  & 1.02 & 1.03 & 1.00 \\
          PART & 1.01 & 1.00 & 1.02 &  & 1.04 & 1.04 & 1.05 &  & 1.05 & 1.09 & 1.03 \\
          NUM & 1.00 & 1.00 & 1.00 &  & 1.00 & 1.00 & 1.00 &  & 1.00 & 1.00 & 1.10 \\
          PUNCT & 1.08 & 1.05 & 1.11 &  & 1.00 & 1.00 & 1.00 &  & - & 1.00 & 1.00 \\
          INTJ & 1.00 & 1.12 & 1.14 &  & 1.00 & 1.05 & 1.17 &  & 1.03 & 1.05 & 1.06 \\
          X & 1.09 & 1.06 & 1.22 &  & 1.00 & 1.08 & 1.00 &  & 2.33 & 1.80 & 2.00 \\
          SYM & - & - & - &  & 1.00 & 1.00 & 1.00 &  & 1.00 & - & - \\
         \bottomrule
    \end{tabular}
    }
    \caption{Context PoS-tag-level raw fertility values for De $\rightarrow$ En, En $\rightarrow$ De, and En $\rightarrow$ Hi language pairs.}
    \label{tab:context_raw_fertility_scores}
\end{table*}


\begin{table*}[!ht]
    \centering
    \resizebox{1.0\linewidth}{!}{
    \begin{tabular}{l||cccc||cccc||ccc}
         \toprule
         \multirow{2}*{\textbf{PoS Tag}} & \multicolumn{3}{c}{\textbf{De $\rightarrow$ En}} & & \multicolumn{3}{c}{\textbf{En $\rightarrow$ De}} & & \multicolumn{3}{c}{\textbf{En $\rightarrow$ Hi}} \\
         \cmidrule{2-12}
          & Prev-1 & Next-1 & Random-1 & & Prev-1 & Next-1 & Random-1 & & Prev-1 & Next-1 & Random-1 \\
          \cmidrule{1-12}
          PRON & 0.07 & 0.06 & 0.16 &  & 0.11 & 0.09 & 0.13 &  & 0.11 & 0.17 & 0.09 \\
          VERB & 0.19 & 0.20 & 0.27 &  & 0.13 & 0.13 & 0.16 &  & 0.37 & 0.21 & 0.20 \\
          ADP & 0.11 & 0.19 & 0.12 &  & 0.10 & 0.13 & 0.06 &  & 0.20 & 0.23 & 0.20 \\
          DET & 0.11 & 0.13 & 0.15 &  & 0.08 & 0.10 & 0.07 &  & 0.17 & 0.15 & 0.22 \\
          ADJ & 0.31 & 0.17 & 0.22 &  & 0.16 & 0.22 & 0.21 &  & 0.22 & 0.25 & 0.11 \\
          NOUN  & 0.19 & 0.17 & 0.28 &  & 0.10 & 0.14 & 0.13 &  & 0.22 & 0.21 & 0.24 \\
          CCONJ & 0.14 & 0.10 & 0.07 &  & 0.06 & 0.12 & 0.00 &  & 0.00 & 0.22 & 0.00 \\
          ADV & 0.14 & 0.16 & 0.26 &  & 0.15 & 0.13 & 0.11 &  & 0.31 & 0.25 & 0.23 \\
          AUX & 0.09 & 0.14 & 0.21 &  & 0.13 & 0.17 & 0.17 &  & 0.28 & 0.22 & 0.24 \\
          PROPN & 0.18 & 0.34 & 0.12 &  & 0.00 & 0.00 & 0.59 &  & 0.45 & 0.28 & 0.45 \\
          SCONJ & 0.07 & 0.00 & 0.25 &  & 0.28 & 0.14 & 0.20 &  & 0.15 & 0.18 & 0.00 \\
          PART & 0.08 & 0.00 & 0.16 &  & 0.24 & 0.24 & 0.28 &  & 0.29 & 0.43 & 0.20 \\
          NUM & 0.00 & 0.00 & 0.00 &  & 0.00 & 0.00 & 0.00 &  & 0.00 & 0.00 & 0.47 \\
          PUNCT & 0.26 & 0.29 & 0.50 &  & 0.00 & 0.00 & 0.00 &  & - & 0.00 & 0.00 \\
          INTJ & 0.00 & 0.54 & 0.59 &  & 0.00 & 0.28 & 0.41 &  & 0.21 & 0.29 & 0.31 \\
          X & 0.44 & 0.34 & 0.76 &  & 0.00 & 0.41 & 0.00 &  & 0.92 & 0.72 & 0.00 \\
          SYM & - & - & - &  & 0.00 & 0.00 & 0.00 &  & 0.00 & - & - \\
         \bottomrule
    \end{tabular}
    }
    \caption{Context PoS-tag-level raw entropy values for De $\rightarrow$ En, En $\rightarrow$ De, and En $\rightarrow$ Hi language pairs.}
    \label{tab:context_raw_entropy_scores}
\end{table*}


\begin{table*}[!ht]
    \centering
    \resizebox{1.0\linewidth}{!}{
    \begin{tabular}{l||ccccc||ccccc||cccc}
         \toprule
         \multirow{2}*{\textbf{PoS Tag}} & \multicolumn{4}{c}{\textbf{De $\rightarrow$ En}} & & \multicolumn{4}{c}{\textbf{En $\rightarrow$ De}} & & \multicolumn{4}{c}{\textbf{En $\rightarrow$ Hi}} \\
         \cmidrule{2-15}
          & No Context & Prev-1 & Next-1 & Random-1 & & No Context & Prev-1 & Next-1 & Random-1 & & No Context & Prev-1 & Next-1 & Random-1 \\
          \cmidrule{1-15}
          PRON & 20.72 & 22.86 & 23.47 & 22.38 &  & 12.70 & 15.16 & 15.46 & 14.27 &  & 12.92 & 15.87 & 15.96 & 16.85 \\
          VERB & 5.37 & 7.82 & 7.87 & 7.24 &  & 9.66 & 11.68 & 11.57 & 11.35 &  & 6.59 & 9.12 & 9.43 & 8.73 \\
          ADP & 13.27 & 10.79 & 10.90 & 11.92 &  & 16.20 & 13.93 & 14.10 & 14.85 &  & 18.02 & 15.45 & 15.42 & 16.02 \\
          DET & 15.30 & 14.26 & 14.22 & 14.22 &  & 12.89 & 11.63 & 11.89 & 12.18 &  & 26.48 & 20.23 & 20.77 & 21.62 \\
          ADJ & 4.82 & 5.45 & 4.98 & 4.72 &  & 3.79 & 4.04 & 4.08 & 3.99 &  & 3.65 & 4.91 & 4.44 & 4.42 \\
          NOUN  & 4.84 & 6.63 & 6.13 & 5.63 &  & 13.26 & 14.02 & 13.88 & 13.41 &  & 10.41 & 13.45 & 13.17 & 11.15 \\
          CCONJ & 2.21 & 2.24 & 2.31 & 2.25 &  & 4.47 & 4.37 & 4.42 & 4.61 &  & 0.47 & 0.51 & 0.46 & 0.42 \\
          ADV & 13.89 & 12.25 & 12.18 & 12.87 &  & 7.93 & 7.33 & 7.29 & 7.42 &  & 5.52 & 5.57 & 5.17 & 5.51 \\
          AUX & 7.33 & 7.20 & 7.48 & 7.55 &  & 10.28 & 9.34 & 9.17 & 9.48 &  & 7.99 & 7.45 & 7.39 & 7.35 \\
          PROPN & 0.42 & 0.38 & 0.42 & 0.33 &  & 0.24 & 0.22 & 0.22 & 0.24 &  & 0.76 & 0.67 & 0.76 & 0.61 \\
          SCONJ & 2.21 & 2.19 & 2.24 & 2.45 &  & 2.17 & 2.42 & 2.34 & 2.37 &  & 0.93 & 1.12 & 1.16 & 1.19 \\
          PART & 6.29 & 5.03 & 4.87 & 5.47 &  & 5.05 & 4.49 & 4.24 & 4.56 &  & 4.25 & 3.60 & 3.74 & 3.83 \\
          NUM & 0.27 & 0.33 & 0.32 & 0.25 &  & 0.33 & 0.42 & 0.35 & 0.30 &  & 0.64 & 0.51 & 0.49 & 0.52 \\
          PUNCT & 2.73 & 2.29 & 2.30 & 2.52 &  & 0.32 & 0.33 & 0.36 & 0.39 &  & 0.17 & 0.09 & 0.15 & 0.13 \\
          INTJ & 0.05 & 0.09 & 0.04 & 0.04 &  & 0.63 & 0.53 & 0.59 & 0.53 &  & 1.19 & 1.45 & 1.52 & 1.64 \\
          X & 0.27 & 0.21 & 0.26 & 0.17 &  & 0.06 & 0.10 & 0.05 & 0.04 &  & 0.00 & 0.00 & 0.00 & 0.00 \\
          SYM & - & - & - & - &  & 0.00 & 0.00 & 0.01 & 0.00 &  & 0.00 & 0.00 & 0.00 & 0.00 \\
         \bottomrule
    \end{tabular}
    }
    \caption{Percentage of zero fertility source tokens at PoS-tag-level for De $\rightarrow$ En, En $\rightarrow$ De, and En $\rightarrow$ Hi language pairs.}
    \label{tab:pos_zero_fertility_raw_scores}
\end{table*}

\end{document}